%% file: _main.tex
\input{_constants}
\arxiv

\pdfoutput=1
\documentclass[10pt,twocolumn,letterpaper]{article}
\input{cvpr_header}
\unless\ifarxiv \myexternaldocument{_supplementary} \fi
\begin{document}
\title{PViT-6D: Overclocking Vision Transformers for 6D Pose Estimation with Confidence-Level Prediction and Pose Tokens}
\author{\authorBlock}

\maketitle

\input{00_abstract}
\input{01_intro}
\input{02_related}

\input{03_method}

\input{04_experiments}

\input{10_conclusion}

{\small
    \bibliographystyle{ieeenat_fullname}
    \bibliography{11_references}
}

\ifarxiv \clearpage \appendix \input{12_appendix} \fi

\end{document}

%% file: _constants.tex
\def\paperID{11513}
\def\confName{CVPR}
\def\confYear{2024}

\def\authorBlock{
Sebastian Stapf\textsuperscript{1,2} \qquad
Tobias Bauernfeind\textsuperscript{2} \qquad
Marco Riboldi\textsuperscript{1} \\
\textsuperscript{1} Ludwig-Maximilians-Universität München, \qquad \textsuperscript{2} BMW Group\\
{\tt\small s.stapf@campus.lmu.de, tobias.baeuernfeind@bmw.de, marco.riboldi@physik.lmu.de }
}

\newif\ifreview 
\newif\ifarxiv \newcommand{\arxiv}{\arxivtrue}
\newif\ifcamera 
\newif\ifrebuttal 

%% file: cvpr_header.tex
\ifreview \usepackage[review]{cvpr} \fi
\ifarxiv \usepackage[pagenumbers]{cvpr} \fi
\ifrebuttal \usepackage[rebuttal]{cvpr} \fi
\ifcamera \usepackage{cvpr} \fi

\input{_macros}  

\usepackage{xr-hyper}
\usepackage[table]{xcolor}

\makeatletter
\newcommand*{\addFileDependency}[1]{
  \typeout{(#1)}
  \@addtofilelist{#1}
  \IfFileExists{#1}{}{\typeout{No file #1.}}
}

\makeatother
\newcommand*{\myexternaldocument}[1]{
    \externaldocument{#1}
    \addFileDependency{#1.tex}
    \addFileDependency{#1.aux}
}

\definecolor{cvprblue}{rgb}{0,1,0}
\usepackage[pagebackref,breaklinks,colorlinks,citecolor=cvprblue]{hyperref}
\usepackage[capitalize]{cleveref}
\crefname{section}{Sec.}{Secs.}
\crefname{table}{Table}{Tables}
\crefname{figure}{Fig.}{Figs.}

%% file: _macros.tex
\usepackage{graphicx}	
\usepackage{amsmath}	
\usepackage{amssymb}	
\usepackage{booktabs}
\usepackage{times}
\usepackage{microtype}
\usepackage{epsfig}
\usepackage[table,xcdraw,dvipsnames]{xcolor}
\usepackage{caption}
\usepackage{float}
\usepackage{placeins}
\usepackage{color, colortbl}
\usepackage{stfloats}
\usepackage{enumitem}
\usepackage{tabularx}
\usepackage{xstring}
\usepackage{multirow}
\usepackage{xspace}
\usepackage{url}
\usepackage{subcaption}
\usepackage{xcolor}
\usepackage[hang,flushmargin]{footmisc}

\ifcamera \usepackage[accsupp]{axessibility} \fi

% ...

\newcommand{\nbf}[1]{{\noindent \textbf{#1.}}}

\newcommand{\supp}{supplemental material\xspace}
\ifarxiv \renewcommand{\supp}{appendix\xspace} \fi

\newcommand{\R}[1]{{%
    \textbf{%
        \ifstrequal{#1}{1}{\textcolor{red}{R#1}}{%
        \ifstrequal{#1}{2}{\textcolor{blue}{R#1}}{%
        \ifstrequal{#1}{3}{\textcolor{magenta}{R#1}}{%
        \ifstrequal{#1}{4}{\textcolor{teal}{R#1}}{%
                           \textcolor{cyan}{R#1}%
        }}}}%
    }%
}}

%% file: 00_abstract.tex
\begin{abstract}
    In the current state of 6D pose estimation, top-performing techniques depend on complex
    intermediate correspondences, specialized architectures, and non-end-to-end algorithms.
    In contrast, our research reframes the problem as a straightforward regression
    task by exploring the capabilities of Vision Transformers for direct 6D pose
    estimation through a tailored use of classification tokens.
    We also introduce a simple method for determining pose confidence, which can be readily integrated into most 6D pose estimation frameworks. This involves modifying the transformer architecture by decreasing the number of query elements based on the network's assessment of the scene complexity.
    Our method that we call \textbf{P}ose \textbf{Vi}sion \textbf{T}ransformer or \textbf{PViT-6D}
    provides the benefits of simple implementation and being end-to-end learnable
    while outperforming current state-of-the-art
    methods by $\textbf{+0.3\% \text{ADD(-S)}}$ on Linemod-Occlusion and
    $\textbf{+2.7\% \text{ADD(-S)}}$ on the YCB-V dataset. Moreover, our method enhances both 
    the model's interpretability and
    the reliability of its performance during inference.
\end{abstract}

%% file: 01_intro.tex
\section{Introduction}
\label{sec:intro}
\begin{figure}[htp!]
    \centering
    \includegraphics[width=\linewidth]{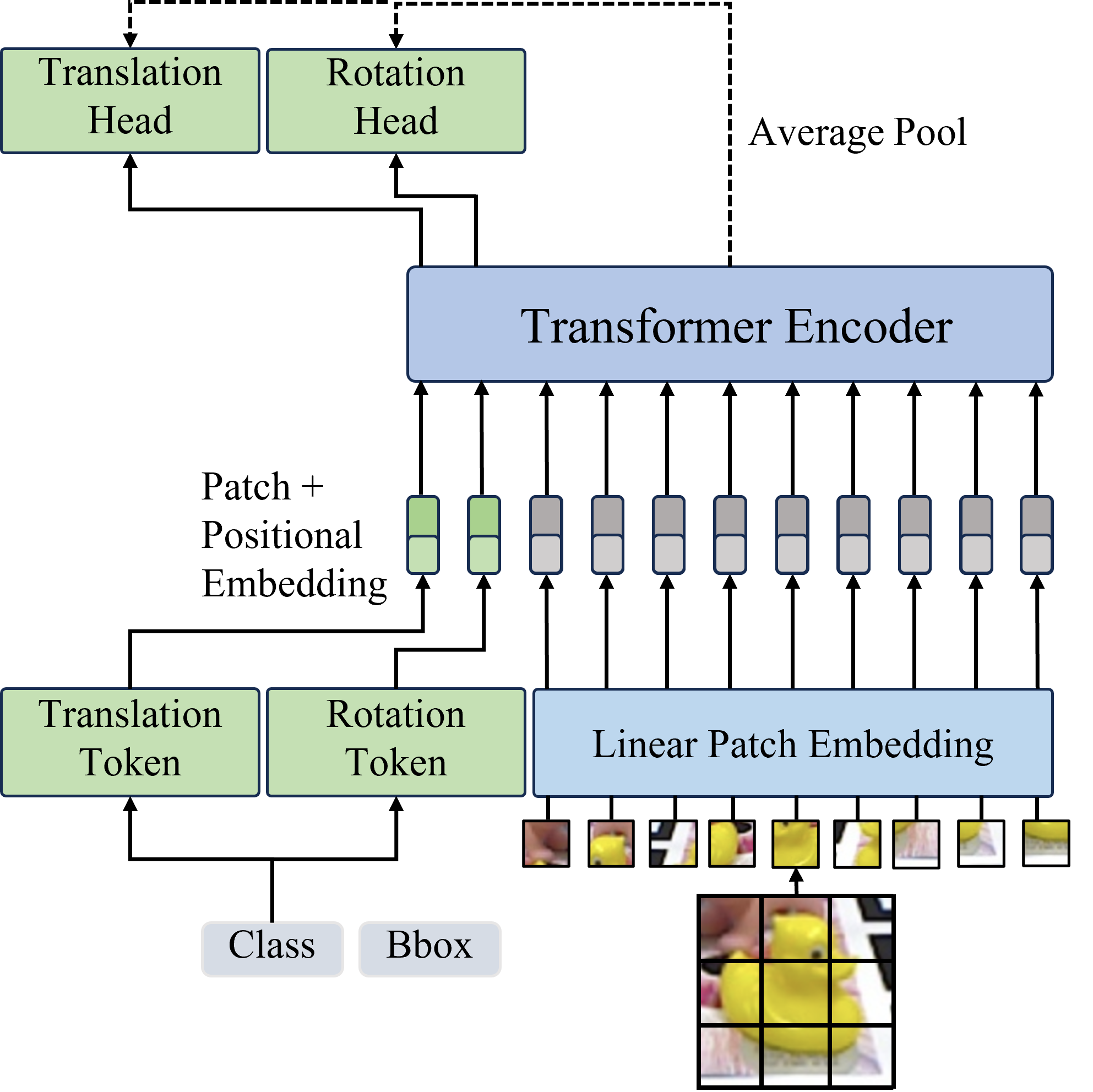}
    \caption{
        Illustration of modifying transformer-based image classification for
        direct pose regression. Rather than relying on a single class token or
        pooling the output features (indicated by the dotted arrow), we
        introduce two extra tokens for predicting rotation and translation.
        Optionally, the model can also be fed additional bounding box information.
    }
    \label{fig:tokens}
\end{figure}
The task of 6D pose estimation presents numerous challenges, mainly
when dealing with scenes that are cluttered or suffer from occlusions. In these
situations, predictive models often generate pose estimations without any
associated measure of reliability. This lack of confidence indicators poses
significant risks in practical applications like autonomous driving \cite{ad1,ad2},
human-machine interfaces \cite{ar1}, and robotics \cite{rob1,rob2,rob3}, where
incorrect predictions can lead to serious consequences.
Solutions that offer confidence metrics often come with restrictions tied to specific
frameworks or types of pose representation, such as keypoint
detection \cite{Tekin2017RealTimeSS,Gupta2019CullNetCA}. These constraints limit the
adaptability of the methods and pose challenges for researchers attempting to incorporate
confidence metrics into their own specialized tasks and models.

Current state-of-the-art techniques in 6D pose estimation have increasingly utilized similar
advanced methods that use intricate intermediate features, such as keypoints
and dense correspondences. This has resulted in significant improvements in accuracy,
as seen in models like PVNet \cite{pvnet}, GDRNet \cite{gdrnet}, and the latest
ZebraPose \cite{zebra}. Typically, these models adopt a simple feature extraction
backbone based on the Resnet architecture \cite{resnet}. This design decision is
influenced by earlier studies, where larger or transformer-based networks were deemed
computationally expensive for both training and real-time applications \cite{vit, gdrnet}.
However, this trade-off comes at the cost of diminished feature quality and predictive
accuracy, leading to ongoing efforts to optimize architecture or employ more complex
representations.

Additionally, these sophisticated techniques introduce their own set of challenges,
such as limited adaptability in multi-modal scenarios and increased computational
and time requirements, particularly when additional ground truth data has to be generated
or real-time inference algorithms need to be optimized \cite{rtdetr, zebra}.

Contrastingly, the foundational work in 6D pose estimation was relatively simple,
mainly relying on direct regression techniques \cite{Bui2018WhenRM, Kehl2017SSD6DMR}.
The shift toward complexity is fueled by the challenging nature of accurately
predicting 3D object poses, which involves both translational and rotational components.
Initial implementations faced numerous issues, including finding the appropriate
representation for rotation and translation, and designing efficient architectures and
loss functions, all of which severely restricted performance and necessitated the
development of more complex methods \cite{gdrnet, Do2018LieNetRM}.

This evolution has made the research landscape in 6D pose estimation distinctly
different from traditional 2D computer vision tasks such as image classification, object
detection, and human pose estimation. In these more established fields, direct
regression-like methods have been quite effective and are generally considered the
norm \cite{vit,deit, Liu2021SwinTV,maxvit,Redmon2015YouOL,Xu2022ViTPoseSV}.

In our investigation, we make a case for applying direct 6D pose regression
within a modern context, incorporating proven pose representations, loss functions,
and cutting-edge transformer-based network designs. We assert that this constitutes
the new standard for fully learnable end-to-end methods, thus bridging the divide
between classical 2D methodologies and the present advances in 6D pose estimation.
Additionally, we demonstrate that linking the 3D Intersection-over-Union (3D-IoU)
metric to the forecasted confidence, provides
a straightforward and effective approach to extend many existing 6D pose estimation algorithms.

%% file: 02_related.tex
\section{Related Work}
\label{sec:related}

\nbf{Direct Pose Estimation}
Posenet \cite{Kendall2015PoseNetAC} pioneered using quaternions for predicting
the rotation in camera pose estimation tasks. This approach brought a new dimension
to the way rotations could be learned and represented.
PoseCNN \cite{posecnn} took a multi-faceted approach by generating vectors that
point towards the object center for each pixel to estimate translation. For rotation,
they opted for direct regression using quaternions, similar to Posenet.
SSD-6D \cite{Kehl2017SSD6DMR} approached the problem by discretizing the viewpoint
space and then classifying it. To estimate the object's distance to the camera, they
employed mask-based regression.
CRT-6D \cite{crt6d} represents a more recent advancement by incorporating
deformable attention modules. This method estimates the object's pose, project keypoints
onto the feature map, and then sample these to refine the initial pose prediction.

\nbf{Indirect Pose Estimation}
PVNet \cite{pvnet} introduced a two-stage method that initially estimates the 2D
keypoints of objects in the image. These keypoints are then used in an
uncertainty-driven Perspective-n-Point (PnP) algorithm to calculate the object's
pose.
GDRNet \cite{gdrnet}, on the other hand, diverged from the use of sparse keypoints
and instead employed dense correspondences. These dense correspondences encode the
relationship between each image pixel and its corresponding point on the 3D model of
the object. In a notable departure from traditional methods, GDRNet substituted the
classical PnP algorithm with a neural network that directly regresses the pose from
these dense correspondences.
ZebraPose \cite{zebra} also adopted a dense correspondence strategy but took a
different route. In this approach, each pixel is classified into a binary class
corresponding to a specific part of the object. This classification is then
leveraged to match the image pixels with points on the object's 3D model.
Subsequently, a PnP algorithm is used to estimate the object's pose.

\nbf{Pose Confidence}
The approach by \citeauthor{Tekin2017RealTimeSS} involves predicting the 2D 
projections of an object's 3D bounding box and computing a score for each point 
based on its distance to the ground truth point. Extending this methodology, 
\citeauthor{Gupta2019CullNetCA} predict 2D keypoints along with their confidence 
scores. These are then used to estimate object poses via an Efficient-PnP algorithm. 
The estimated poses enable the generation of masks and bounding boxes, which 
further refine the confidence scores through a Convolutional Neural Network (CNN) 
at processes both the mask and the Region of Interest (RoI).

%% file: 03_method.tex
\section{Method}
\label{sec:method}
We aim to develop a 6D pose estimation technique that estimates the pose
directly from an image, bypassing intermediary steps. However, in images
with multiple objects, direct regression becomes ambiguous. To tackle this,
we draw inspiration from previous works \cite{gdrnet,crt6d, zebra} and
employ a two-step approach: initial object detection by an object detector,
followed by separate pose estimation for each detected RoI.
After the RoIs are extracted from the object detector,
consistently with the recent studies \cite{crt6d,gdrnet}, we
scale the RoIs by a factor of 1.2, crop them out, and
resize the cropped image to a fixed size of 224x224 before using them as input
for the network's first layer.

Furthermore, we aim to separate the tasks of regressing the rotation and
translation components of the pose. We specifically
tailor the pose representation, loss functions, and network architecture to accomplish this.

\subsection{Pose Representation}
In the task of 6D pose estimation, particularly the prediction of the rotation
$\mathbf{R}\in \text{SO}(3)$ and translation $\mathbf{t}\in\mathbb{R}^3$ using neural networks, the selection
of an appropriate pose representation is a critical factor affecting
the model's performance \cite{representations}.
Recent studies have shown the advantages of using
a 6D rotation representation, denoted as \( \mathbf{R}_{6\mathrm{d}} \), in
excelling at 6D pose estimation tasks \cite{gdrnet,crt6d,poet}.
The 6D rotation representation $\mathbf{R}_{6 \mathrm{d}}$ is defined as
\begin{equation}
    \mathbf{R}_{6 \mathrm{d}}=\left[\mathbf{r}_{1} \mid \mathbf{r}_{2}\right] .
\end{equation}
where  $\mathbf{r}_{1}, \mathbf{r}_{2} \in \mathbb{R}^{3}$ are the first two column vectors of the rotation matrix.
The rotation matrix
$\mathbf{R} $ can be recovered according to Gram-Schmidt
orthonormalization. For rotational representation, we use the object-centric allocentric
representation due to its equivariance with the RoI's appearance \cite{3drcnn}.

Handling 3D translation in the camera's coordinates is challenging because
it is not linearly associated with pixel changes. To solve this, we adopt
the method from \citeauthor{posecnn}, which decomposes translation into
2D centroid coordinates $\left(c_{x}, c_{y}\right)$ and the object's
distance $t_{z}$ from the camera, simplifying 3D translation calculation
via back-projection:
\begin{equation}
    \mathbf{t} = K^{-1} t_{z} \left[ c_{x}, c_{y}, 1 \right]^{T},
    \label{eq:backprojection}
\end{equation}
where $K$ denotes the camera's intrinsic matrix.

Given that object detectors already supply an approximate 2D location of
the object through the predicted bounding box center
$\left(o_{x}, o_{y}\right)$ and the network is only given the RoI to infer the pose,
we employ the Scale-Invariant Translation
Parameterization (SITP) for further refinement \cite{cdpn}. Specifically, we predict the offset
between the object's centroid and the bounding box center, normalized by the
size of the bounding box. The SITP is then defined as
$\mathbf{t}_{\mathrm{SITP}}=\left[\gamma_{x}, \gamma_{y}, \gamma_{z}\right]^{T}$,
where
\begin{equation}
    \label{eq:site}
    \left\{\begin{array}{l}
        \gamma_{x}=\left(c_{x}-o_{x}\right) / s \\
        \gamma_{y}=\left(c_{y}-o_{y}\right) / s \\
        \gamma_{z}=t_{z} / r
    \end{array} .\right.
\end{equation}
Here, $s=\max(w, h)$ represents the size of the bounding box, and $r$ is the
ratio of the image size $s_{\text{inp}}$ to the bounding box size, defined
as $r = s_{\text{inp}}/s$.

However, the above representation is scale-agnostic, which means
that the model cannot distinguish between a small object that is close to
the camera and a large object that is far away. To address this issue,
we introduce the Size Invariant Z Parametrization (SIZP). We utilize the
back-projection formula (\ref{eq:backprojection}) to derive:
\begin{equation}
    t_z = \frac{f_{\text{s}}\cdot s_\text{obj}}{s}
\end{equation}
where $f_{\text{s}}$ is the focal length of the camera, and $s_{\text{obj}}$
is the real-world size of the object along the $s$ direction. Notice
that this approach is only viable if the object detector provides us with
the class of the object.
While the real-world size of the object, $s_{\text{obj}}$, is dependent
on the object's dimensions and is not precisely known due to rotation, we
can approximate it using the object's diameter $d$ and the fraction of the
object's diameter visible in the bounding box, denoted as $\gamma_z$.
Consequently, the variable that our model aims to predict is formulated as
\begin{equation}
    \label{eq:sizp}
    \gamma_{z}^{(\text{class})}=\frac{t_{z}}{r}=\frac{t_z\cdot s}{f_s\cdot d^{(\text{class})}}
\end{equation}
As an extension of SITP, this representation provides a size-invariant description
of the object's translation. As demonstrated in our experiments, another benefit
of this representation is that it scales the predicted variables for the
translation to be in a similar range, which enhances the training stability
without the need for further tuning.

\subsection{Pose Tokens}
\label{sec:tokens}
For image classification tasks with vision transformers, \citeauthor{vit} introduced
the concept of class tokens, which are learnable embeddings
that are appended to the feature map before feeding it into the transformer
to aggregate context information for the classification task.
For our tasks the class token provides two benefits. First by initializing
the token with a learned embedding of the class from the object detector,
we can aggregate class specific features. Second, we can use multiple tokens
for different prediction tasks.
\begin{figure}[htp]
    \centering
    \includegraphics[width=1.0\linewidth]{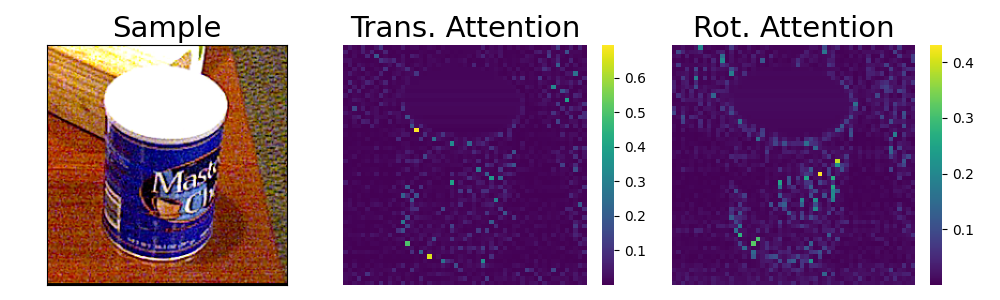}
    \caption{Visualization of the attention maps for a handpicked sample.
        Left the input image, in the middle the attention map of the translation
        token, and right the rotation token. To predict the rotation of the symmetric object, the rotation token needs to capture
        the label of the can.}
    \label{fig:token_attention}
\end{figure}
Hence  we further tailor the architecture to make it more conducive for 6D pose
estimation tasks. We add two additional tokens to the network: one for
aggregating the object's translation and another for predicting its rotation features, see \cref{fig:tokens}.
The idea is to allow the model to learn different features for the rotation and
translation prediction, as depicted by an example in \cref{fig:token_attention}.
Lastly for predicting the pose, we use the pose tokens as input to two separate 3-layer Multi-Layer-Perceptrons (MLPs) to predict the
rotation and translation individually.

\subsection{Pose Confidence}
\label{sec:pose_confidence}
\begin{figure*}[htp]
    \centering
    \hspace*{\fill} 

    \begin{subfigure}{0.225\textwidth}
        \includegraphics[width=\linewidth]{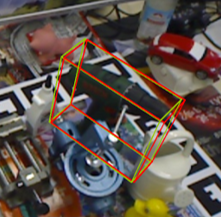}
        \caption*{\textbf{Pred. Score: 0.90\\ GT Score: 0.89}}
    \end{subfigure}
    \hfill
    \begin{subfigure}{0.225\textwidth}
        \includegraphics[width=\linewidth]{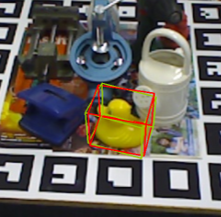}
        \caption*{\textbf{Pred. Score: 0.82\\ GT Score: 0.73}}
    \end{subfigure}
    \hfill
    \begin{subfigure}{0.225\textwidth}
        \includegraphics[width=\linewidth]{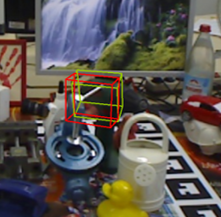}
        \caption*{\textbf{Pred. Score: 0.28\\ GT Score: 0.14}}
    \end{subfigure}
    \hfill
    \begin{subfigure}{0.225\textwidth}
        \includegraphics[width=\linewidth]{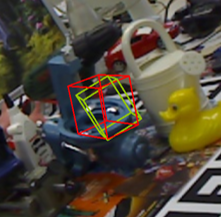}
        \caption*{\textbf{Pred. Score: 0.39\\ GT Score: 0.54}}
    \end{subfigure}

    \caption{Visualization of predicted confidence score and ground truth 3D-IoU score.
        The green 3D bounding box depicts the box transformed with the ground truth pose, while the red box is transformed with the predicted pose.
        Below the imges the individual predicted scores and ground truth scores are shown.}
    \label{fig:score3d}
\end{figure*}

Motivated by the contributions of \citeauthor{Zhang2020VarifocalNetAI,Tekin2017RealTimeSS},
we consider a method for assessing the reliability of the pose
predictions.
To obtain a ground
truth score that our model aims to predict, we employ both the ground truth
pose and the estimated pose to transform the 3D bounding box of the object and calculate the 3D-IoU.
Even though straightforward, as shown by RT-SSD6D \cite{Tekin2017RealTimeSS}, the method becomes infeasible for dense object
detectors or pose estimators due to the usually high number of predictions
$\geq 100$ and the
computational demand of the 3D-IoU \cite{detr}. However, the method gets enabled in the
2-stage setting since we need to calculate only one 3D-IoU per object detection.
The 3D-IoU is then formulated as:
\begin{equation}
    \text{IoU}_{3D} = \frac{\text{Volume of Intersection}}{\text{Volume of Union}}.
\end{equation}
As illustrated in \cref{fig:score3d}, the ground truth score quantifies
the degree to which the predicted pose aligns with the actual ground truth
pose.

However rather than predicting a single scalar score, we estimate a class-specific score, defined as:
\begin{equation}
    \textbf{GT-Score}(c,p,\hat{p}) = \hat{c}_{\text{scale}}\left(\text{IoU}_{3D}(c,\hat{p},p)\right) \cdot \textbf{OneHot}(c),
\end{equation}
where \( c \) denotes the object class, \( p \) and \( \hat{p} \) represent
the ground truth and estimated poses, respectively, and \( \hat{c}_{\text{scale}} \)
serves as a scaling operator, treated as a hyperparameter, that scales the dimensions of the bounding box if
the average translation error is much bigger than the original size of the
objects.
This method enhances class-specific detection, crucial for class-dependent representations like SIZP, see \cref{eq:sizp}.

Lastly, our two-stage approach faces the issue of training on ground truth RoIs and classes, not accounting for object detector errors. We address this by using a denoising method inspired by \citeauthor{dino}, which involves injecting noise through false positives, either by changing the class randomly or removing the object from the RoI. This trains the model to handle detection inaccuracies.

\subsection{Scene Complexity Conditioned Attention}
\label{sec:attention}



\begin{figure}[t]
    \centering
    \includegraphics[width=\linewidth]{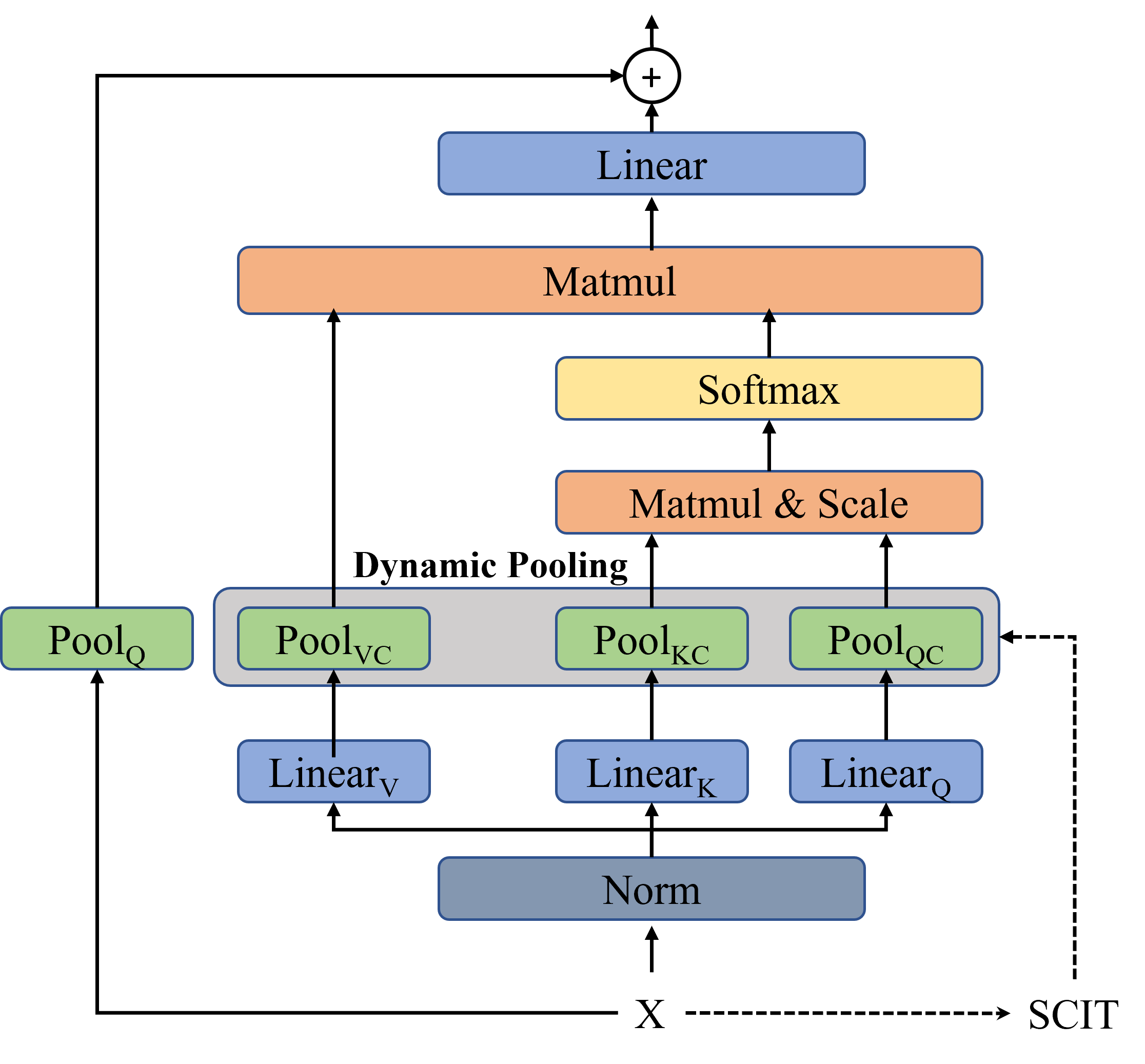}
    \caption{Depiction of the Scene-Complexity-Conditioned-Attention Mechanism. The SCIT is used as a condition for the pooling operations to extract the relevant features from the feature map, dependent on the scene context.}
    \label{fig:block}
\end{figure}

We base our architecture on the Multi-Scale-Vision-Transformer (MViT) \cite{mvit, mvitv2} that provides key improvements such as multiscale strategies adopted from classical CNNs and reduced sequence
lengths via pooled attention. We further adapt the MViT architecture to our specific application by introducing a Scene-Complexity-Conditioned-Attention (SCCA) mechanism.

To account for variations in pose estimation accuracy caused by scene complexity and the challenges in predicting rotation and translation, we introduce the Scene Complexity Identifier Token (SCIT), denoted as \(C_{\text{SCIT}} \in \mathbb{R}^{D}\). We design this token to condense relevant features from the feature map according to the scene's context, helping filter out extraneous or misleading feature elements by using it to predict the pose confidence.

We formalize our attention mechanism as follows:
Consider an input feature map \(X \in \mathbb{R}^{H \times W \times D}\), and define the linear projections \(W_Q, W_K, W_V \in \mathbb{R}^{D \times D}\). The feature map is then transformed by the pooling operators \(\mathcal{P}_Q\), \(\mathcal{P}_K\), and \(\mathcal{P}_V\), which are applied after the projections as:
\begin{align*}
    \tilde{Q} = \mathcal{P}_Q(XW_Q), \
    \tilde{K} = \mathcal{P}_K(XW_K), \
    \tilde{V} = \mathcal{P}_V(XW_V),
\end{align*}
yielding reduced-dimensionality maps \(\tilde{Q}, \tilde{K}, \tilde{V} \in \mathbb{R}^{\tilde{H} \times \tilde{W} \times D}\) with \(\tilde{H} \leq H\) and \(\tilde{W} \leq W\), as illustrated in \cref{fig:block}.

\begin{figure}[ht]
    \centering
    \begin{subfigure}{0.49\linewidth}
        \centering
        \includegraphics[scale=0.55]{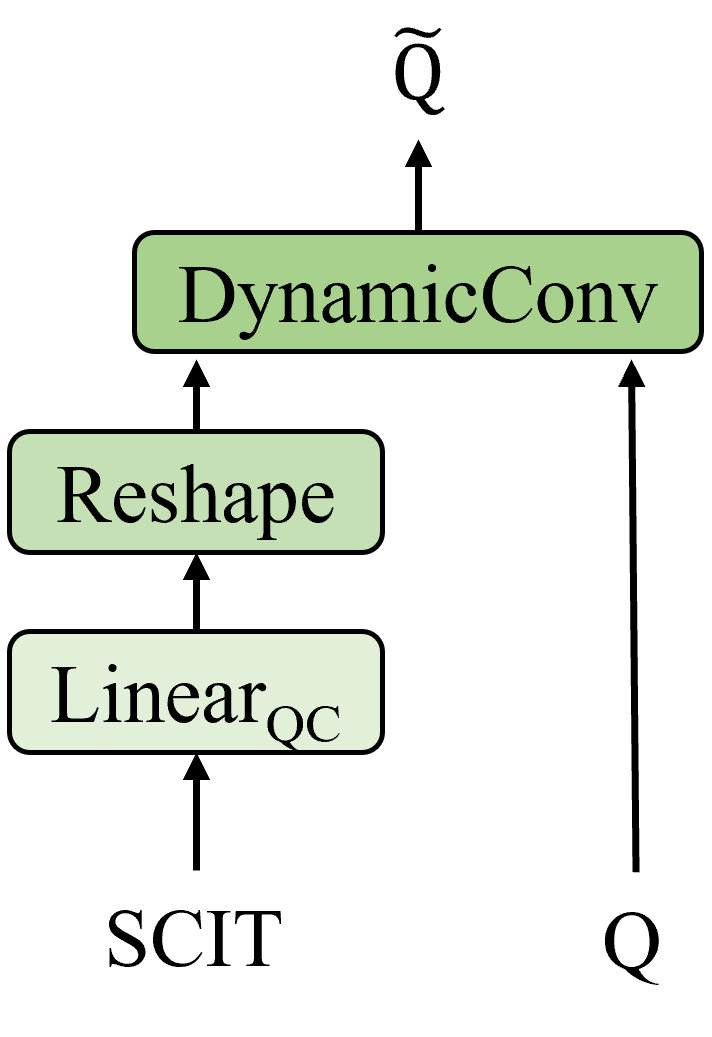}
        \caption{Dynamic-Filter}
        \label{fig:sub1}
    \end{subfigure}
    \begin{subfigure}{0.49\linewidth}
        \centering
        \includegraphics[scale=0.55]{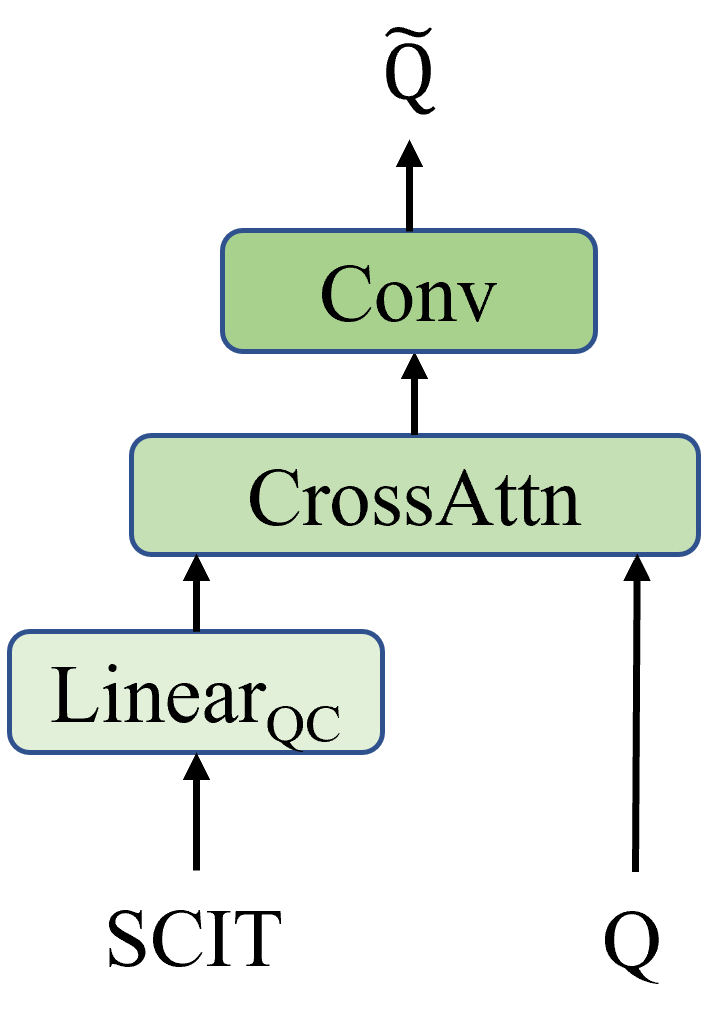}
        \caption{Cross-Attention}
        \label{fig:sub2}
    \end{subfigure}
    \caption{Scene-Complexity-Pooling Module: (a) Dynamic Convolution based, the SCIT is used to generate weights for the Convolution, (b) right
        cross-attention based pooling, the SCIT is used to generate attention maps, which are then used to pool the feature map with a learned convolution.}
    \label{fig:pool}
\end{figure}

We also integrate a standard convolutional pooling operation, \(\mathcal{P}_{Q}\), serving as a residual connection for the input feature map.
As depicted in \cref{fig:pool}, the pooling operations are conditioned on the SCIT vector in two distinct methodologies. Firstly, we apply a dynamic filter convolution \cite{Jia2016DynamicFN} defined by:
$\mathcal{K}_Q = \text{Linear}_{Q_C}(C_{\text{SCIT}}),$
where \(\mathcal{K}_Q \in \mathbb{R}^{D \times D \times k_h \times k_w}\). The corresponding conditional pooling operator, \(\mathcal{P}_{Q_C}\), is:
\begin{align*}
    \mathcal{P}_{Q_C}(Q) & = \text{Conv}_{\mathbf{s, p}}(Q, \mathcal{K}_Q) = \tilde{Q},
\end{align*}
with \(Q = XW_Q \in \mathbb{R}^{H \times W \times D}\), and stride and padding parameters denoted by \(\mathbf{s} =(s_h, s_w)\) and \(\mathbf{p} = (p_h, p_w)\), respectively.

Secondly, we utilize a cross-attention mechanism \cite{Vaswani2017AttentionIA}.
During the forward pass, attention weights are computed using
\begin{align*}
    A = \text{Softmax}(\sum_{i=1}^{D}\left(\text{Linear}_{Q_C}(C_{{\text{SCIT}}})_i \cdot Q_i\right) \cdot \alpha),
\end{align*}
with \( \alpha \) being a scaling factor. The attended feature map, \( Q_{\text{attended}} = Q \cdot A \), is processed to obtain \( \tilde{Q} = \text{Conv}_{\mathbf{s,p}}(Q_{\text{attended}}) \) with learned weights.
And similar of course for $\mathcal{P}_{K_C}$ and $\mathcal{P}_{V_C}$.
Ultimately, the attention mechanism is calculated as:
\begin{align*}
    Z & = \text{Attn}(Q, K, V) = \text{Softmax}\left( \frac{\tilde{Q}\tilde{K}^T}{\sqrt{D}} \right)\tilde{V}.
\end{align*}

Lastly we use three different models with scales, Small, Base, and Large, which we denote as PViT-6D-s, PViT-6D-b, and PViT-6D-l.
For simplicity and comparability, we use the same model configurations as in MViTv2 \cite{mvitv2}.

\subsection{Loss Function}
In addressing the pose loss, we adhere to the same loss function that
has been employed in numerous recent works\cite{gdrnet,crt6d,sopose,posecnn}, which have demonstrated
notable success.
Drawing inspiration from \citeauthor{disentangle}, we decouple the rotation
and translation loss. Consequently, the pose loss is defined
as follows:
$\mathcal{L}_{\text{Pose}} = \lambda_\text{c}\mathcal{L}_{\text{centroid}} +\lambda_z\mathcal{L}_{z} + \lambda_\mathbf{R}\mathcal{L}_{\mathbf{R}}$,
where
\begin{align}\label{eq:your_label}
    \left\{\begin{array}{l}
               \mathcal{L}_{\text{center}} = \left\|\left(\hat{\gamma}_{x} - {\gamma}_{x}, \hat{\gamma}_{y} - {\gamma}_{y}\right)\right\|_{1}, \\
               \mathcal{L}_{z} = \left\|\hat{\gamma}_{z} - {\gamma}_{z}\right\|_{1},                                                           \\
               \mathcal{L}_{\mathbf{R}} = \underset{\mathbf{x} \in \mathcal{M}}{\operatorname{avg}}\left\|\hat{\mathbf{R}} \mathbf{x} - {\mathbf{R}} \mathbf{x}\right\|_{1}
           \end{array}\right.
\end{align}
Here $\left|\left| \cdot \right|\right|_1$ denotes the $L1$-norm and $\mathcal{M}$ is the model of the predicted object.
The $\lambda$-s are used as hyperparameters to weight the different losses.
The loss for the rotation is a variant of the pose matching loss \cite{posecnn},
where only the rotation is used to transform the points. For objects with symmetrical
characteristics, we employ a symmetry-aware pose loss as used by \citeauthor{gdrnet}
This loss is defined as
$\min_{\mathbf{R}\in\mathcal{R}} \mathcal{L}_\mathbf{R}(\hat{\mathbf{R}}, \mathbf{R})$.
Here, the loss function takes into account the object's inherent symmetries to
determine the rotation closest to the ground truth rotation. In this equation,
$\mathcal{R}$ represents the set of all feasible ground truth rotations that conform
to the object's symmetry.

\noindent
\noindent
For the class loss we opt for a simple cross entropy loss:
\begin{align}
    \mathcal{L}_{\text{cls}}(c,\hat{s}, p, \hat{p}) & = \mathcal{L}_{\text{cls}}(c,\hat{s}, \text{IoU}_{3d}) \\
                                                    & =- \sum_{i=1}^{N} \textbf{GT-Score}_i \log(\hat{s}_i)
\end{align}

%% file: 04_experiments.tex
\section{Experiments}
\label{sec:experiments}
In this section, we delve into the specifics of the experiments.
We begin by outlining the datasets and evaluation
criteria employed. Subsequently, we conduct ablation studies.
We conclude by comparing our findings
with those obtained using state-of-the-art approaches on both the
LM-O \cite{lmo} and YCB-V \cite{posecnn} datasets and investigate the
effectiveness of the pose confidence.

\subsection{Datasets}
The Linemod (LM) \cite{lm} dataset comprises 15 scenes, each with around 1,200 images and a single annotated object, providing ground truths like 3D models, bounding boxes, masks, and 6D object poses \cite{lm}. Due to high accuracy achieved by various methods \cite{posecnn, zebra, gdrnet, cdpn}, research has pivoted to the more challenging LM-O dataset \cite{lmo}, which adds a test set with 1,214 images and an average of eight occluded objects per image. We leverage the LM and the BOP challenge's PBR datasets, the latter containing 50 scenes with approximately 1,200 rendered images each \cite{bop}, to enhance pose estimation model performance \cite{gdrnet, hybridpose}.

Furthermore, our experiments extend to the YCB-V dataset \cite{posecnn}, with 21 objects across 92 video-recorded scenes, presenting heavier occlusions and intricate symmetries, backed by over 100,000 real images and additional PBR data from the BOP challenge \cite{bop} for training.

\subsection{Evaluation Metrics}
For the evaluation of the pose estimation, we employ the commonly used ADD(-S) metric
\cite{evalpose}.
The ADD(-S) metric consists of the ADD metric,
which is defined to be the
average distance between the 3D points after transforming them with the predicted
and ground truth pose. For symmetric objects, we use the ADD-S metric, which is
defined to be the minimal distance between the points after transforming them.
For LM-O, we use the average
recall (AR) of the ADD(-S) metric, where a true pose is given when the prediction
has an ADD(-S) score below a certain threshold. Other works \cite{zebra,gdrnet}
have used a threshold of 10\% of the object's diameter,
which we also use in our experiments.

Additionally, for the YCB-V dataset, we also report the area under the curve (AUC)
of the ADD(-S) metric, where the threshold is varied in the interval
    [0, 10cm] \cite{posecnn}.

\subsection{Implementation Details}
\label{sec:implementation}

The implementation of the model is done using PyTorch \cite{pytorch}, and it
is trained across three Nvidia 3080Ti GPUs. We employ the AdamW optimizer \cite{adamw}
with a learning rate set to 1.2e-4 and a batch size configured
to 192. We use a linear learning rate scheduler, incorporating a 2-epoch
warm-up phase. Additionally, we set the weight decay to 1e-4, and the gradient
norms are clipped at 0.1. The training lasts 150 epochs, with
early stopping with a patience set to 15. For
image augmentation, we apply random adjustments in brightness, contrast, saturation,
hue, coarse dropout, and color jitter. In terms of bounding box augmentation,
we perform uniform scaling and shifting by 10\% and 35\% of the original
size, respectively.

\subsection{Ablation Study}
To test changes, we implement an ablation dataset. We use the LM dataset with an additional ten scenes from the PBR dataset to test on the LM-O dataset. We call this dataset LM-O-A.

\nbf{Influence of z-representation}

To test the absolute z-representation, we incorporate a bounding box embedding with
the token embeddings. Our findings reveal the
model's difficulty in accurately predicting the absolute z-coordinate with \textbf{54.6\% ADD(-S)}
accuracy, while the relative z-coordinate(SITP) and our proposed representation(SIZP) achieve, in comparison,
higher accuracies of \textbf{+6.5\%} and \textbf{+7.2\%}, respectively.

\nbf{Influence of Class Refinement}
Further, we investigate the influence of the class refinement. In a test scenario
where we disturb 20\% of the ground-truth classes, the refinement model recovers
the correct class in \textbf{92.9\%}, and with our denoising strategy
\textbf{94.7\%} of the cases. We compare
the model's performance with and without class refinement, utilizing
the SIZP representation. We see a slight increase of \textbf{+0.2\% ADD(-S)} accuracy when
using Faster-RCNN detections from \citeauthor{Ren2015FasterRT}.

\nbf{Dependency on Encoder Architecture}
We want to investigate the influence of the encoder architecture
on the performance of the model. We use
a standard vision transformer as a baseline ViT-B/16 \cite{vit} and multiple
versions of the MViTv2 \cite{mvitv2} architecture. MViTv2-s, MViTv2-b, MViTv2-l,
with varying depth and parameter count.
We also add one convolutional-based backbone ResNet-50 \cite{resnet} for a comparison,
where we average pool the output feature map.
\begin{table}[ht]
    \centering
    \begin{tabular}{l|c|c|c}
        \rowcolor{lightgray}
        \textbf{Model}              & \textbf{ADD(-S)}        & \textbf{GFLOPs}         & \textbf{Param (M)}       \\
        \hline
        ResNet-50*                  & 43.1                    & 4.3                     & 25.6                     \\
        ViT-B/16                    & 62.7                    & 17.7                    & 86.4                     \\
        MViTv2-s                    & 56.9                    & 7.1                     & 29.8                     \\
        \textcolor{black}{MViTv2-b} & \textcolor{black}{62.0} & \textcolor{black}{10.4} & \textcolor{black}{44.9}  \\
        \textcolor{black}{MViTv2-l} & \textcolor{black}{65.2} & \textcolor{black}{42.9} & \textcolor{black}{201.2} \\
    \end{tabular}
    \caption{\textbf{Dependency on the Encoder Architecture.} Depicted are Encoder-Architecture, GLFOPS and parameter count in million. * means no tokens are used, ADD(-S) accuracy reported on LM-O-A.}
    \label{tab:backbone}
\end{table}
The results depicted in \cref{tab:backbone} show, that the MViTv2 architecture scales well with the number
of parameters and provides a better speed-accuracy tradeoff than the
ViT architecture, by only having -0.7\% accuracy with only 52\% the amount of parameters.

\nbf{Pose Token Ablation}
To test the effectiveness of using two tokens for translation
and rotation, we also experiment with using a single pose token. To have a fair
comparison with the pool baseline, we concatenate the pooled output feature
vector with a learned class embedding.
We find that using a single pose token for both translation and rotation
already achieves an improvement of \textbf{+3.2\%} over the pool baseline with 57.4\% accuracy compared to 60.6\%.
Additionally, segregating pose tokens for translation and rotation enhances
performance by an additional \textbf{+1.4\%} over using a single token, suggesting the model's ability to
distill distinct features for each.

\nbf{Influence of SCCA}
For comparing the effectiveness of the SCCA from \cref{sec:attention}, we use the base pose token version,
with and without the cross-attention- and the convolution-based SCCA.
Furthermore, we investigate the effects of using the conditioned pooling
only on the queries.
\begin{table}[ht!]
    \centering
    \begin{tabular}{l|c|c|c}
        \rowcolor{lightgray} 
        \textbf{Variant}                     & \textbf{ADD(-S)}                 & $\Delta$\textbf{GFLOPs} & \textbf{Param (M)}      \\
        \hline
        None                                 & 62.0                             & \bfseries{10.3}         & \bfseries{51.8}         \\
        {Conv$^q$}                           & -0.2                             & +0.1                    & 53.8                    \\
        \bfseries\textcolor{black}{Attn$^q$} & \bfseries\textcolor{black}{+0.5} & \textcolor{black}{+0.2} & \textcolor{black}{52.2} \\
        {Conv$^{q,k,v}$}                     & +0.0                             & +0.4                    & 57.7                    \\
        {Attn$^{q,k,v}$}                     & +0.4                             & +1.0                    & 53.1                    \\
    \end{tabular}
    \caption{\textbf{Comparison of SCCA-versions}. GFLOPs are reported as difference from the
        Pool-version. ADD(-S) accuracy reported on the LM-O-A dataset}
    \label{tab:scop}
\end{table}
\cref{tab:scop} indicates that while the convolutional-based SCCA does not enhance
model performance despite more parameters, the attention-based SCCA improves it by
\textbf{+0.5\%} with minimal parameter and computational increase. Additionally,
applying conditioned pooling solely to queries is more effective than on keys and
values in terms of accuracy and computational efficiency.

\subsection{Comparison with State of the Art}
For evaluating LM-O, we employ FasterRCNN detections, which are publicly
accessible \cite{Ren2015FasterRT} and also conduct evaluations using the
FCOS \cite{Tian2019FCOSFC} detections supplied by \citeauthor{zebra}.
We observe an increase of \textbf{+1.1\%} in accuracy (from 71.2\% using PViT-6D-b) by going from
FasterRCNN to FCOS detections, which is in line with the results of
\citeauthor{zebra}.
For assessments on the YCB-V dataset, we leverage the FCOS detections made
available by CDPNv2 \cite{cdpn}.

As shown in \cref{tab:sota}, our smallest model PViT-6D-s model is on par with
methods like GDRNet \cite{gdrnet}, SOPose \cite{sopose}.
Moreover, we outperform the current state-of-the-art  pose estimation model
ZebraPose \cite{zebra} by \textbf{+0.3\%} on the LM-O dataset, and \textbf{+3.3\%} on the YCB-V dataset,
with our largest version, PViT-6D-l, utilizing the same FCOS detections.
Notably, PViT-6D-l, despite having over seven times more parameters than ZebraPose,
effectively infers poses for all classes. On YCB-V, PViT-6D-l shows a
\textbf{+2.7\%} improvement over ZebraPose, and excels in symmetric objects
accuracy by an average of \textbf{+19\%}, as detailed in \supp.

\begin{table*}[ht!]
    \centering
    \begin{tabular}{lcccccccc}
        \toprule
        Model       & PVNet \cite{pvnet} & GDRNet \cite{gdrnet} & SO-Pose \cite{sopose} & CRT-6D\cite{crt6d} & {ZebraPose} \cite{zebra} & \text{\textcolor{black}{Ours(-s)}} & \text{\textcolor{black}{Ours(-b)}} & \textbf{\textcolor{black}{Ours(-l)}} \\
        N           & 8                  & 8                    & 1                     & 1                  & 8                        & {\textcolor{black}{1}}             & {\textcolor{black}{1}}             & \textbf{\textcolor{black}{1}}        \\
        Params(M)   & -                  & 28.8                 & -                     & -                  & {28.9}                   & {\textcolor{black}{31.8}}          & {\textcolor{black}{52.2}}          & \textbf{\textcolor{black}{213.6}}    \\
        \midrule
        Ape         & 15.8               & 46.8                 & 48.4                  & 53.4               & {57.9}                   & {\textcolor{black}{43.3}}          & {\textcolor{black}{ 58.6}}         & \textcolor{black}{ \textbf{60.6}}    \\
        Can         & 63.3               & 90.8                 & 85.8                  & 92.0               & {95.0}                   & {\textcolor{black}{91.8}}          & {\textcolor{black}{ 96.0}}         & \textcolor{black}{ \textbf{97.4}}    \\
        Cat         & 16.7               & 40.5                 & 32.7                  & 42.0               & \textbf{60.6}            & {\textcolor{black}{44.1}}          & {\textcolor{black}{ 56.4}}         & \textcolor{black}{ {60.0}}           \\
        Driller     & 65.7               & 82.6                 & 77.4                  & 81.4               & {94.8}                   & {\textcolor{black}{88.8}}          & {\textcolor{black}{ 93.3}}         & \textcolor{black}{ \textbf{95.8}}    \\
        Duck        & 25.2               & 46.9                 & 48.9                  & 44.9               & {64.5}                   & {\textcolor{black}{56.5}}          & {\textcolor{black}{ 58.3}}         & \textcolor{black}{ \textbf{66.0}}    \\
        Eggbox*     & 50.2               & 54.2                 & 52.4                  & 62.7               & \textbf{70.9}            & {\textcolor{black}{47.5}}          & {\textcolor{black}{ 62.3}}         & \textcolor{black}{68.4}              \\
        Glue*       & 49.6               & 75.8                 & 78.3                  & 80.2               & \textbf{88.7}            & {\textcolor{black}{75.4}}          & {\textcolor{black}{ 84.6}}         & \textcolor{black}{85.9 }             \\
        Holepuncher & 36.1               & 60.1                 & 75.3                  & 74.3               & {83.0}                   & {\textcolor{black}{57.0}}          & {\textcolor{black}{ 75.5}}         & \textcolor{black}{\textbf{83.9} }    \\
        \midrule
        Average     & 40.8               & 62.2                 & 62.3                  & 66.3               & 76.9                     & {\textcolor{black} {63.1}}         & {\textcolor{black} {72.3}}         & \textbf{\textcolor{black} {77.3}}    \\
        \bottomrule                                                                                                                                                                                                                                      \\
    \end{tabular}
    \caption{\textbf{Comparison with the State of the Art on LM-O.} We report the Average Recall (\%) of ADD(-S).
        N is the number pose estimators for the dataset. In the case of LM-O $N_{\text{max}}=8$. ($\ast$) denotes symmetric objects and ``-'' denotes unavailable results.}
    \label{tab:sota}
\end{table*}

\begin{table}[h]
    \centering
    \begin{tabular}{|c|c| p{1.35cm} p{1.35cm}|}
        \hline
        Method                                & N                              & \centering ADD(-S)                          & {AUC of \newline ADD(-S)} \\
        \hline
        PVNet\cite{pvnet}                     & 21                             & \centering -                                & 73.4                      \\
        GDRNet\cite{gdrnet}                   & 21                             & \centering 60.1                             & 84.4                      \\
        SO-Pose\cite{sopose}                  & 1                              & \centering 56.8                             & 83.9                      \\
        CRT-6D\cite{crt6d}                    & 1                              & \centering  72.1                            & \bfseries87.5             \\
        Zebra\cite{zebra}                     & 21                             & \centering 80.5                             & 85.3                      \\
        \hline
        \textcolor{black}{Ours(-b)}           & \textcolor{black}1             & \centering\textcolor{black}{77.3}           & \textcolor{black}{83.7}   \\
        \textcolor{black}{\bfseries Ours(-l)} & \textcolor{black}{\bfseries 1} & \centering\textcolor{black}{ \bfseries83.2} & \textcolor{black}{85.6}   \\
        \hline
    \end{tabular}
    \caption{\textbf{Comparison with State of the Art on YCB-V.} Reported are Average Recall of ADD(-S) (\%)
        and the AUC of the ADD(-S).}
    \label{tab:ycbv}
\end{table}

\subsection{Confidence Analysis}
\label{sec:confidence}
\begin{figure}[ht]
    \centering
    \includegraphics[width=\linewidth]{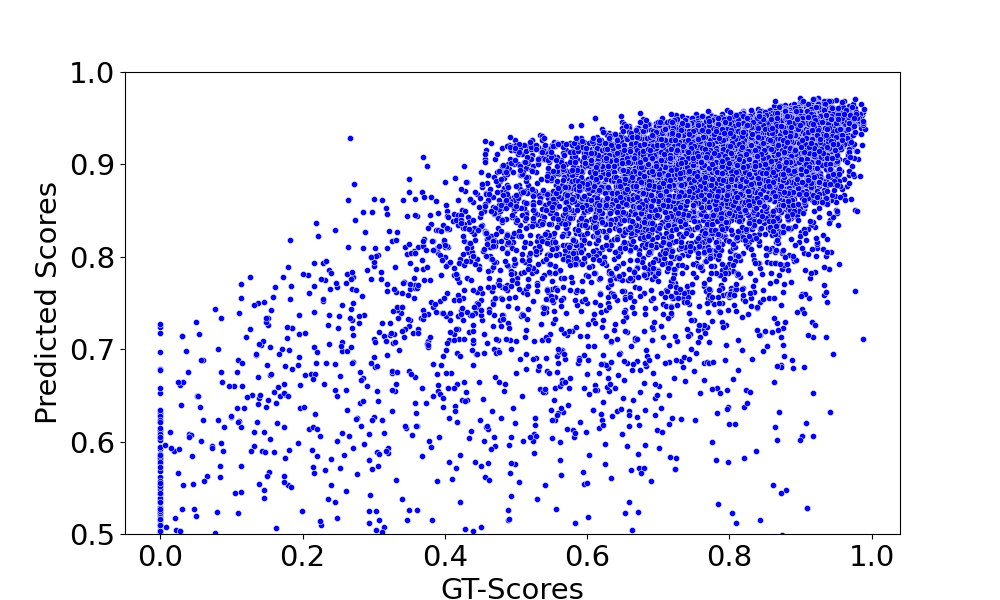}
    \caption{Predicted classification scores and ground truth IoU scores visualized in
        a scatter plot with classification scores of \>0.5, resulting in a
        Spearman's coefficient $\rho=0.71$.}
    \label{fig:score3}
\end{figure}

To evaluate the effectiveness of the pose confidence from \cref{sec:pose_confidence}, we visualize both
predicted confidence scores and ground truth 3D-IoU scores in \cref{fig:score3}. We notice
a pattern of concentration around higher IoU scores, consistent with
our ADD(-S) value of 72.3\% with PViT-6D-b, see \cref{tab:sota}. However, the scores exhibit a significant
degree of variance, indicating the challenges in predicting them accurately.
It is crucial to clarify that the objective is not to predict the 3D-IoU score
precisely.
Instead, the aim is to establish a relationship between
poses that are easier to predict and those that are more challenging.

\begin{figure}[ht]
    \centering
    \includegraphics[width=\linewidth]{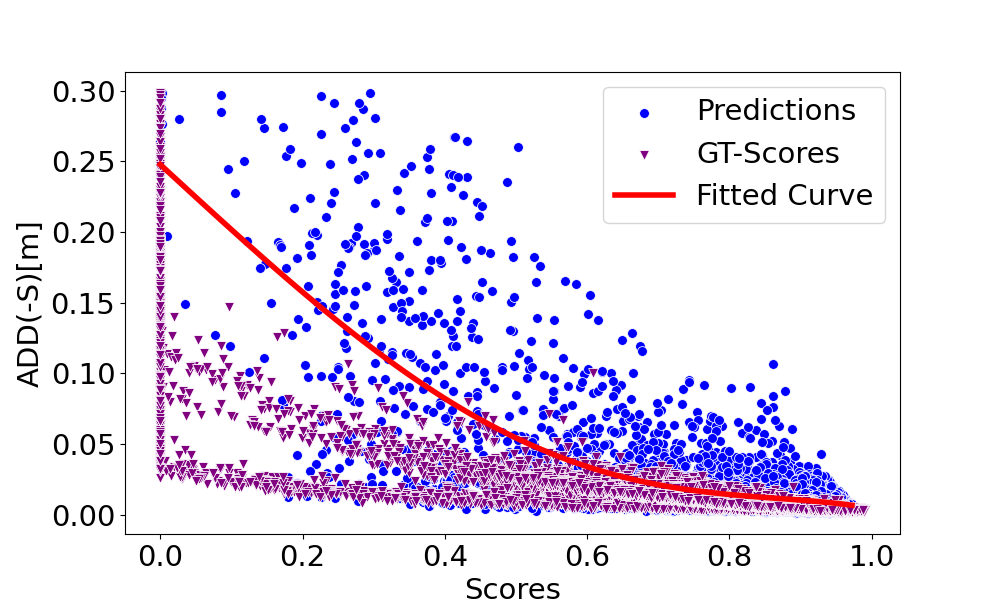}
    \caption{Scatter plot of ADD(-S) error in meters and scores. Shown are the \textcolor{black}{ground
            truth classification scores} and \textcolor{black}{predicted scores} in a scatter plot. Additionally
        shown is \textcolor{black}{polynomial fit of the predicted scores} of degree 10.}
    \label{fig:score1}
\end{figure}

To further analyze the effectiveness of the pose confidence, we visualize the
predicted confidence and ground truth 3D-IoU scores with respect to the ADD(-S) metric
in meters, see \cref{fig:score1}, resulting in a steady decrease in pose error.
\begin{figure}[ht!]
    \centering
    \includegraphics[width=\linewidth]{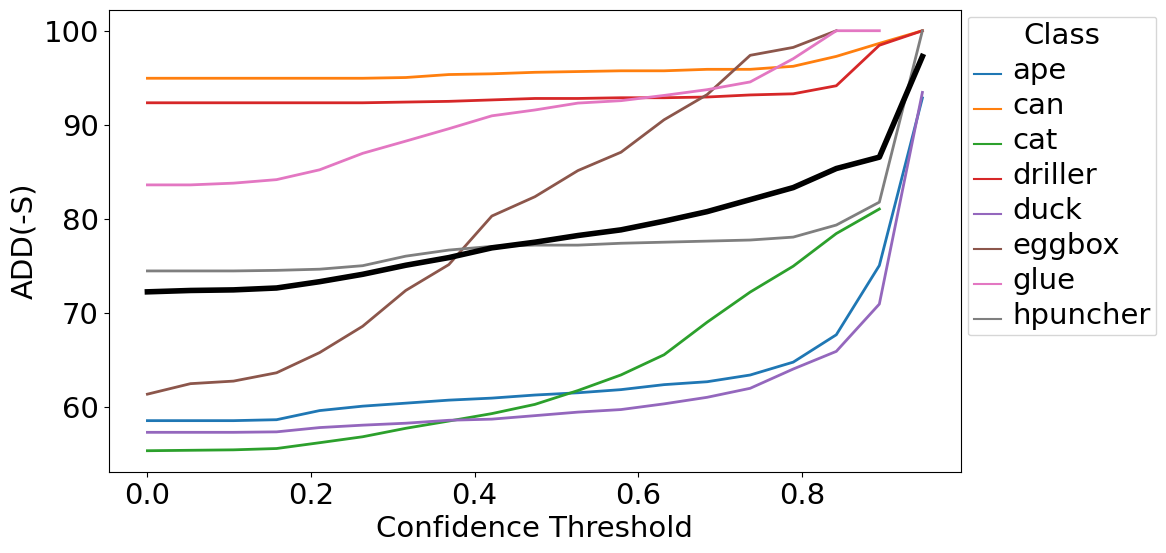}
    \caption{Plotted curves of  predicted confidence scores and ADD(-S) accuracy for each class. In black is the mean curve.}
    \label{fig:score2}
\end{figure}
\textcolor{black}{
    In \cref{fig:score2}, we explore the relationship between class,
    confidence levels, and the accuracy of ADD(-S) measurements.
    We observe a gradual ascent in precision for specific objects due to inherent
    high precision for these objects and the only few incorrect low-confidence
    predictions, which aligns with our expectations.
    There is a notable upward trend in accuracy corresponding to the
    application of stricter confidence thresholds. This trend and the beforementioned results suggests that
    pose confidence is a reliable predictor of pose estimation accuracy.
}

\subsection{Runtime Analysis}

The assessment was conducted on a system equipped with a GTX1080Ti
and an Intel i7-6850K CPU @ 3.60GHz. We evaluated the accuracy and
the inference latency for different object counts and model sizes.
The CRT-6D value is taken from \cite{crt6d} and was measured on a similar setup.

\begin{figure}[ht]
    \centering
    \includegraphics[width=0.9\linewidth]{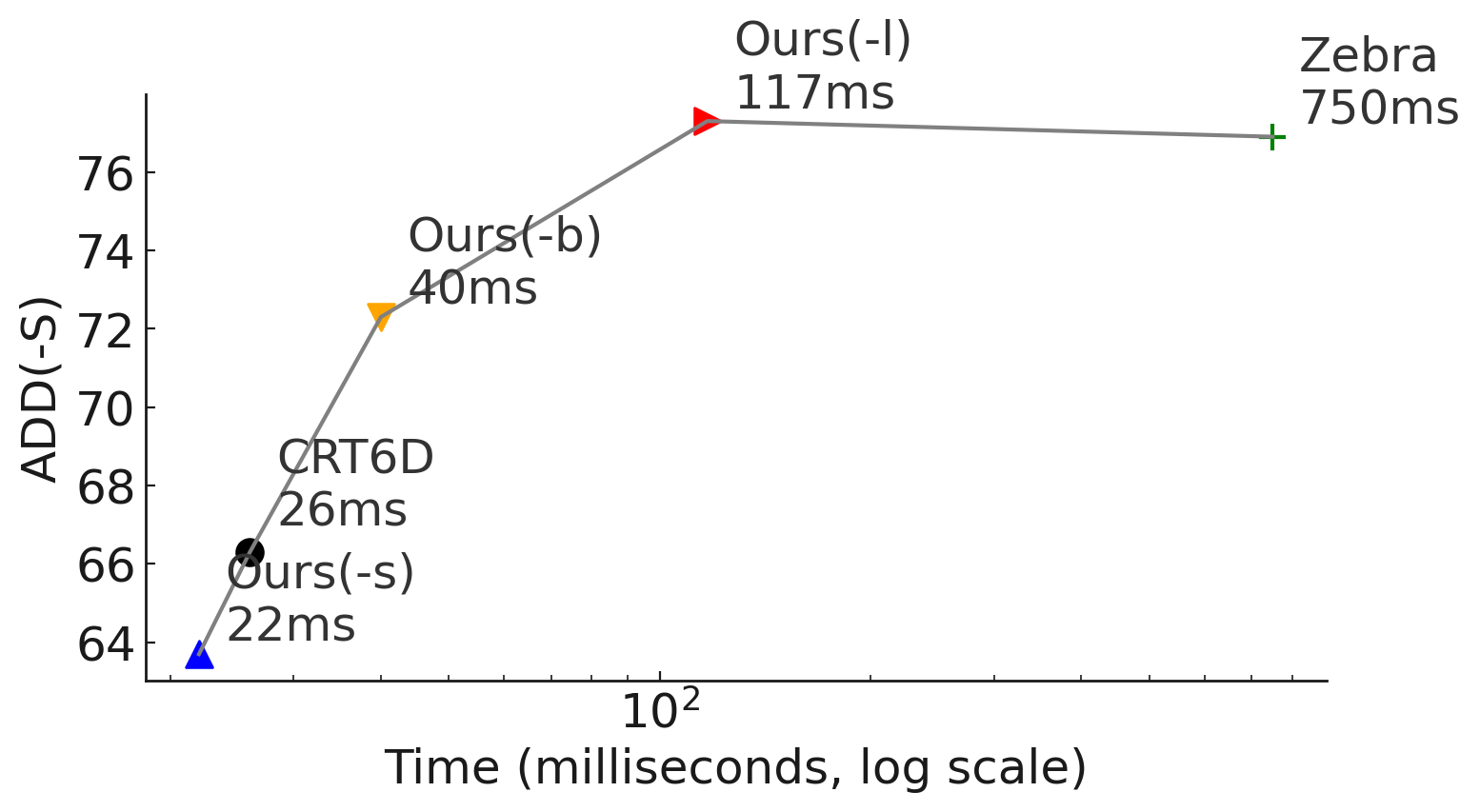}
    \caption{Inference Time Analysis on YBC-V with $\sim4$ objects, excluding object detection. Shown is the ADD(-S) against the inference time in [ms].}
    \label{fig:speed}
\end{figure}
PViT-6D-s exhibits slight advantage in speed while outperformed by CRT-6D\cite{crt6d}
in accuracy.
Zebrapose\cite{zebra}, in our
configured environment, required 188ms to process a single object.
So our largest model
not only outperforms the state-of-the-art in accuracy but also in speed.

%% file: 10_conclusion.tex
\section{Conclusion}
\label{sec:conclusion}
    In conclusion, in our study, we demonstrated the viability of using direct regression for
    pose estimation by using specialized pose
    tokens, yielding both robust
    results and high performance. Additionally, we succeeded in learning
    a measure of pose confidence that is an effective indicator of the model's certainty of a prediction.
    For future research, we aim to implement this method 
    in a multi-modal setting, where pose estimation
    is a subset of the model's capability.

\newpage

%% file: 12_appendix.tex

\section{Influence of Padding Information}
To evaluate whether we can reduce the amount of redundant information we feed into
the model, we investigate 
an alternative to squaring the bounding box, which might potentially introduce
confusing information by showing parts of other objects in occluded scenes.
Instead of squaring, we crop the image
precisely to the bounding box and pad it into a square shape using zero padding.
As it turns out, on LM-O-A, see \cref{sec:experiments}, there is no benefit
in this approach, resulting in a decrease in average recall of ADD(-S) by 0.1\%.


\section{BOP Results}
We train PViT-6D-b/l, under the BOP-setup \cite{bop} on the
YCB-V \cite{posecnn} and LM-O \cite{lmo} datasets and compare them to the previously examined 
state-of-the-art methods and
the two overall top-performing methods, GDRNPP and GPose, based on GDRNet \cite{gdrnet}, and
ZebraSAT, based on ZebraPose \cite{zebra} on the \href{https://bop.felk.cvut.cz/leaderboards/}{BOP-Leaderboard}
with the highest core average recall for RGB-only prediction. Additionally we
take the best-performing method on YCB-V and LM-O individually, namely
CosyPose \cite{cosy} and RADet building on the work of \citeauthor{rad, rad1}.
We use the detections
from CDPNv2 \cite{cdpn} for comparability. Unlike before, now we use only PBR data for training
on LM-O and evaluate only on a set of selected keyframes.
We uploaded the predictions to the \href{https://bop.felk.cvut.cz}{BOP-Website},
where the results are evaluated and submitted under the name \textbf{PViT-6D}.
We show in \cref{tab:bop_lmo} and \cref{tab:bop_ycbv} the results competitive results of 
our method on LM-O and YCB-V, respectively.
\begin{table}[htp!]
    \centering
    \begin{tabular}{lcccc}
        \toprule
        Method                      & \textbf{$AR_{MSPD}$}     & \textbf{$AR_{MSSD}$}            & \textbf{$AR_{VSD}$}             & \textbf{$AR$}                   \\
        \midrule
        CRT-6D \cite{crt6d}         & 83.7                     & 64.0                            & 50.4                            & 66.0                            \\
        Zebra \cite{zebra}          & 88.0                     & 72.1                            & 55.2                            & 71.8                            \\
        GDRNPP                      & 88.7                     & 70.1                            & 54.9                            & 71.3                            \\
        ZebraSAT                    & \textbf{88.9}            & 73.6                            & 56.2                            & 72.9                            \\
        GPose                       & 87.9                     & 68.3                            & {53.6}                          & 69.9                            \\
        RADet                       & 87.7                     & \textbf{76.1}                   & \textbf{59.7}                   & \textbf{74.5}                   \\
        \midrule
        Ours(-b)                    & {87.7}                   & {75.3}                          & {58.8}                          & {73.0}                          \\
        \textcolor{blue}{Ours(-l)*} & \textcolor{blue}{{88.8}} & \textcolor{blue}{\textbf{79.2}} & \textcolor{blue}{\textbf{62.0}} & \textcolor{blue}{\textbf{76.7}} \\
        \bottomrule
    \end{tabular}
    \caption{\textbf{Results on the LM-O dataset under BOP setup \cite{bop}.} GDRNPP,
        ZebraSAT, GPose are the current overall top-performing methods in the BOP-challenge.
        And RADet the top-performing method on LM-O.
        The results  are obtained from the \href{https://bop.felk.cvut.cz/leaderboards/}{BOP-Leaderboard}.
        (*) PViT-6D-l has been trained on PBR and real data.}
    \label{tab:bop_lmo}
\end{table}
\begin{table}[htp!]
    \centering
    \begin{tabular}{lcccc}
        \toprule
        Method              & \textbf{$AR_{MSPD}$} & \textbf{$AR_{MSSD}$} & \textbf{$AR_{VSD}$} & \textbf{$AR$} \\
        \midrule
        CRT-6D \cite{crt6d} & 77.4                 & 77.6                 & 70.6                & 75.2          \\
        Zebra \cite{zebra}  & {86.4}               & 83.0                 & 75.1                & 81.5          \\
        GDRNPP              & 86.9                 & {84.6}               & {76.0}              & {82.5}        \\
        ZebraSAT            & \textbf{87.2}        & 84.2                 & 77.0                & 83.0          \\
        GPose               & 85.3                 & 83.0                 & \textbf{78.3}       & 82.4          \\
        CosyPose            & \textbf{88.0}        & \textbf{88.5}        & \textbf{79.5}       & \textbf{85.3} \\
        \midrule
        Ours(-b)            & 84.0                 & 84.4                 & 75.4                & 81.3          \\
        Ours(-l)            & 85.7                 & \textbf{87.1}        & 76.9                & \textbf{83.2} \\
        \bottomrule
    \end{tabular}
    \caption{\textbf{Results on the YCB-V dataset under the BOP-setup \cite{bop}.} GDRNPP,
        ZebraSAT, GPose are the current overall top-performing methods in the BOP-challenge and
        CosyPose \cite{cosy}, the best-performing method on YCB-V.
        The results  are obtained from the \href{https://bop.felk.cvut.cz/leaderboards/}{BOP-Leaderboard}.}
    \label{tab:bop_ycbv}
\end{table}

\section{Detailed Results on YCB-V}
\label{sec:app_ycb}
In \cref{tab:det_ycbv}, we present the detailed results on the YCB-V dataset \cite{posecnn},
w.r.t. ADD(-S) metric. We compare our method to other state-of-the-art methods.
We show that our method PViT-6D-l outperforms the current state-of-the-art method ZebraPose \cite{zebra} in terms of average recall of ADD(-s).
\begin{table*}[p]
    \centering
    \begin{tabular}{|c|c|c|c|c|c|c|}
        \hline
        Object                        & Single-Stage \cite{Hu2019SingleStage6O} & GDR-Net \cite{gdrnet} & CRT6D \cite{crt6d} & ZebraPose \cite{zebra}   & Ours(-b)  & Ours(-l)                             \\
        \hline
        002\_master\_chef\_can       & -                                      & 41.5                 & -                 & \textbf{62.6}           & 27.4  & 28.7                             \\
        003\_cracker\_box            & -                                      & 83.2                 & -                 & \textbf{95.5}           & 84.3  & 90.3                             \\
        004\_sugar\_box              & -                                      & 91.5                 & -                 & 96.3                    & 94.8  & \textbf{100.0}                   \\
        005\_tomato\_soup\_can       & -                                      & 65.9                 & -                 & 80.5                    & 89.3  & \textbf{94.9}                    \\
        006\_mustard\_bottle         & -                                      & 90.2                 & -                 & \textbf{100.0}          & 95.7  & 98.9                             \\
        007\_tuna\_fish\_can         & -                                      & 44.2                 & -                 & \textbf{70.5}           & 69.2  & 70.4                             \\
        008\_pudding\_box            & -                                      & 2.8                  & -                 & \textbf{99.5}           & 84.6  & 85.1                             \\
        009\_gelatin\_box            & -                                      & 61.7                 & -                 & \textbf{97.2}           & 67.3  & 79.0                             \\
        010\_potted\_meat\_can       & -                                      & 64.9                 & -                 & \textbf{76.9}           & 70.2  & 69.5                             \\
        011\_banana                  & -                                      & 64.1                 & -                 & 71.2                    & 78.7  & \textbf{92.6}                    \\
        019\_pitcher\_base           & -                                      & 99.0                 & -                 & 100.0                   & 100.0 & 100.0                            \\
        021\_bleach\_cleanser        & -                                      & 73.8                 & -                 & 75.9                    & 81.2  & \textbf{88.1}                    \\
        \textcolor{black}{024\_bowl*}& -                                      & 37.7                 & -                 & \textcolor{black}{18.5} & 85.5  & \textcolor{black}{\textbf{99.5}} \\
        025\_mug                     & -                                      & 61.5                 & -                 & 77.5                    & 81.0  & \textbf{93.9}                    \\
        035\_power\_drill            & -                                      & 78.5                 & -                 & \textbf{97.4}           & 94.7  & 97.2                             \\
        036\_wood\_block*            & -                                      & 59.5                 & -                 & 87.6                    & 72.9  & \textbf{99.9}                   \\
        037\_scissors                & -                                      & 3.9                  & -                 & \textbf{71.8}           & 43.8  & 56.3                             \\
        040\_large\_marker           & -                                      & 7.4                  & -                 & 23.3                    & 21.1  & \textbf{27.0}                    \\
        051\_large\_clamp*           & -                                      & 69.8                 & -                 & 87.6                    & 97.3  & \textbf{98.3}                    \\
        052\_extra\_large\_clamp*    & -                                      & 90.0                 & -                 & \textbf{98.0}           & 92.5  & 93.8                             \\
        061\_foam\_brick*            & -                                      & 71.9                 & -                 & 99.3                    & 99.0  & \textbf{99.6}                    \\
        \hline
        Average                      & 53.9                                   & 60.1                 & 72.5              & 80.5                    & 77.3      & \textbf{83.2}                             \\
        \hline
    \end{tabular}
    \caption{\textbf{Detailed Results on YCB-V:} Reported is the Average Recall of ADD(-S) in \%.
        (*) denotes symmetric objects, (-) denotes the results are not reported in the literature.}
    \label{tab:det_ycbv}
\end{table*}

\section{Visual Examples of Attention Maps}
\label{sec:app_attn}

\begin{figure*}[ht]
    \begin{subfigure}{0.48\textwidth}
        \centering
        \includegraphics[width=\linewidth]{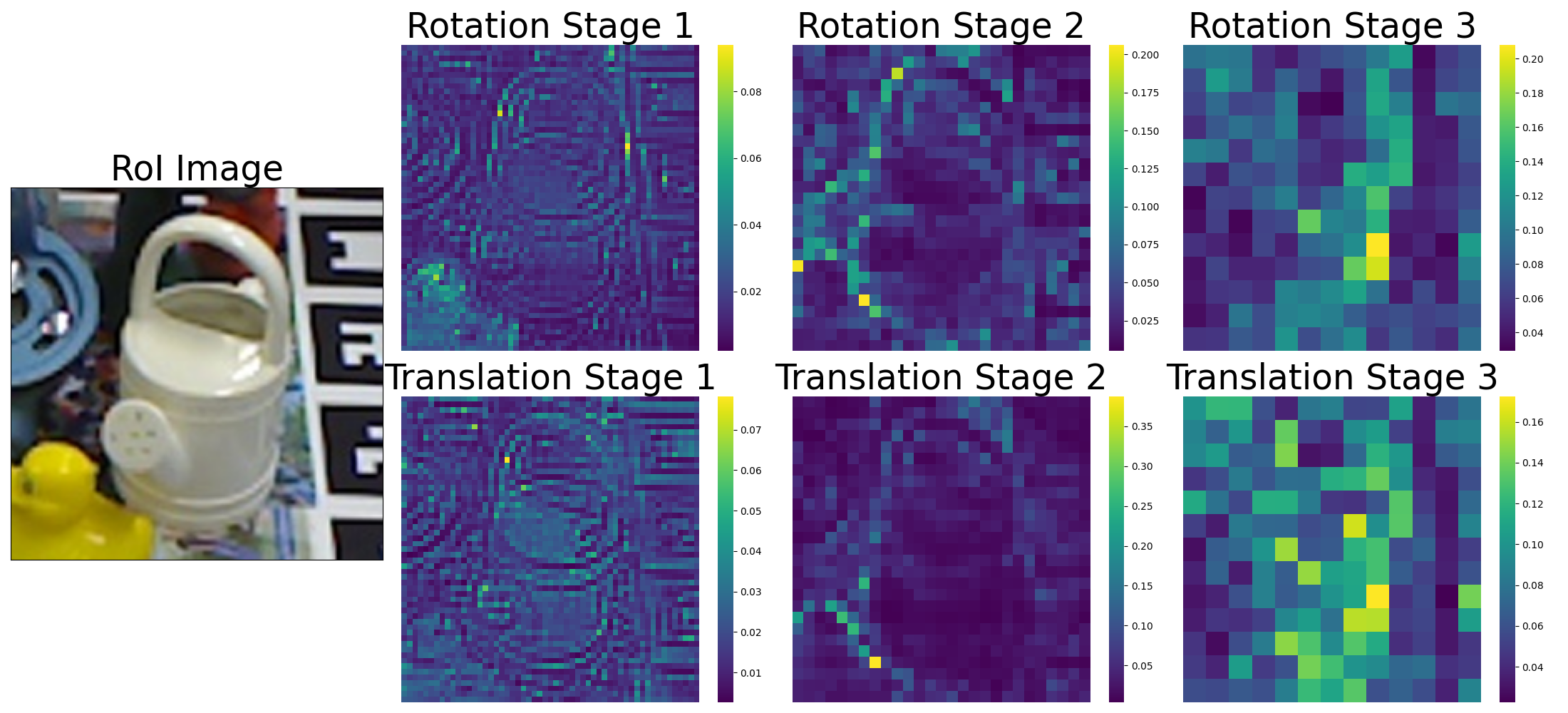}
    \end{subfigure}
    \hfill
    \begin{subfigure}{0.48\textwidth}
        \centering
        \includegraphics[width=\linewidth]{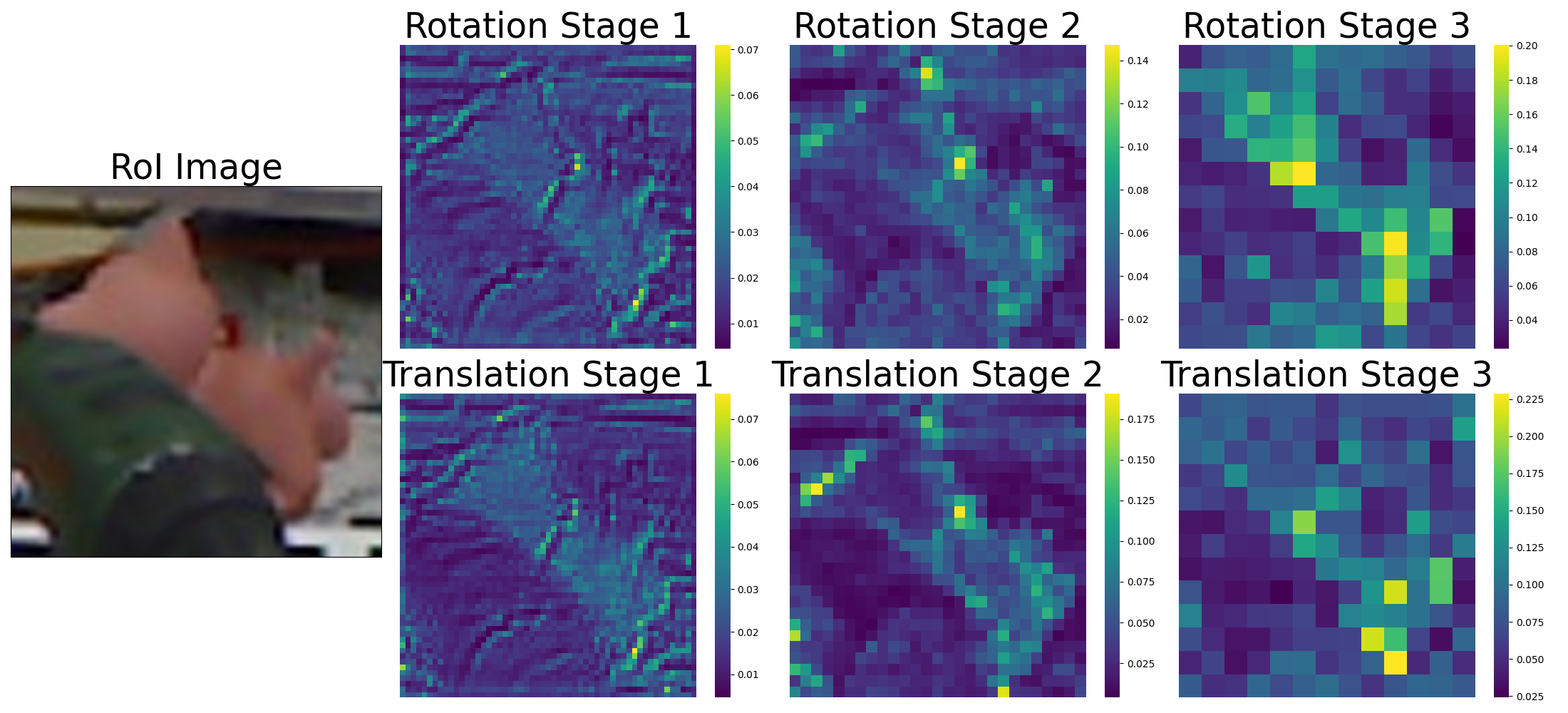}
    \end{subfigure}
    \vspace{10mm}

    \begin{subfigure}{0.48\textwidth}
        \centering
        \includegraphics[width=\linewidth]{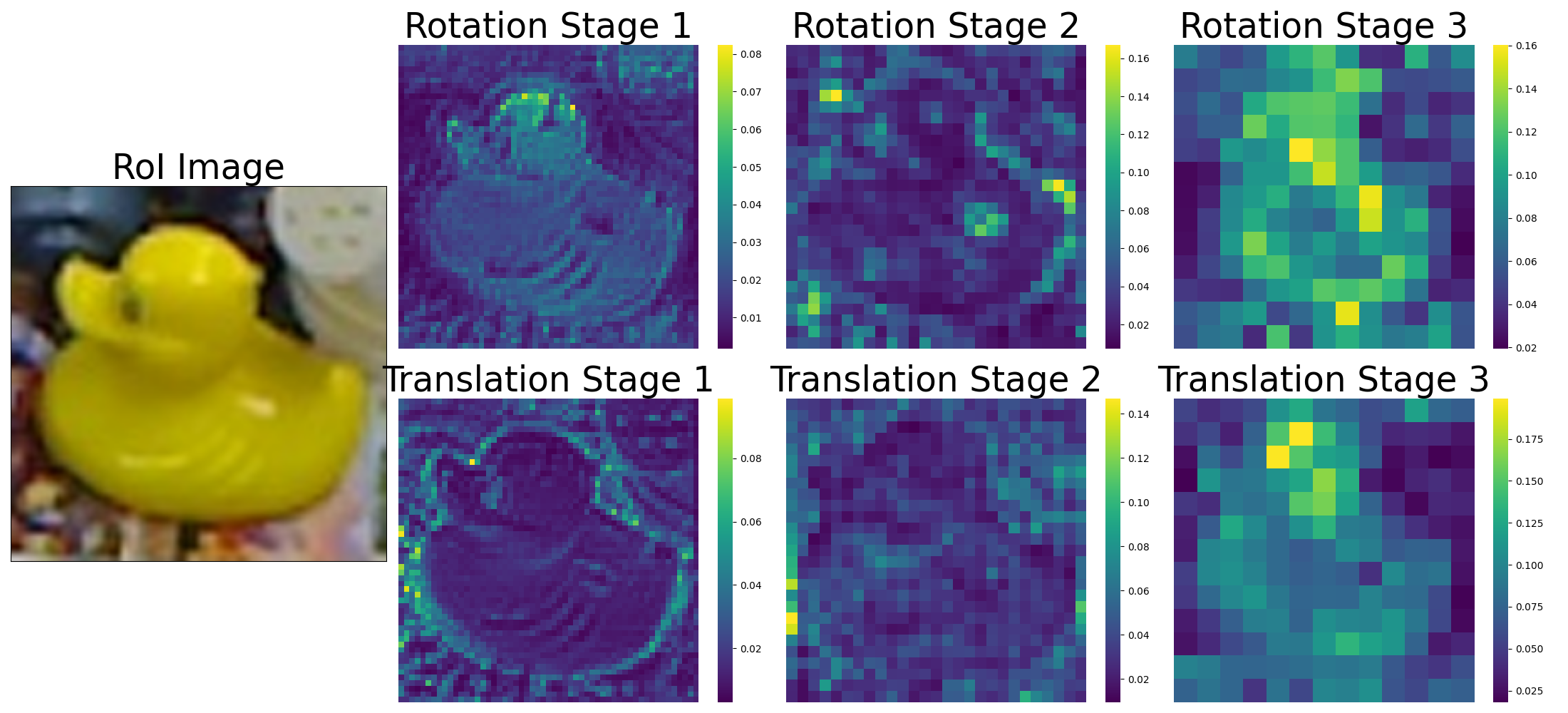}
    \end{subfigure}
    \hfill
    \begin{subfigure}{0.48\textwidth}
        \centering
        \includegraphics[width=\linewidth]{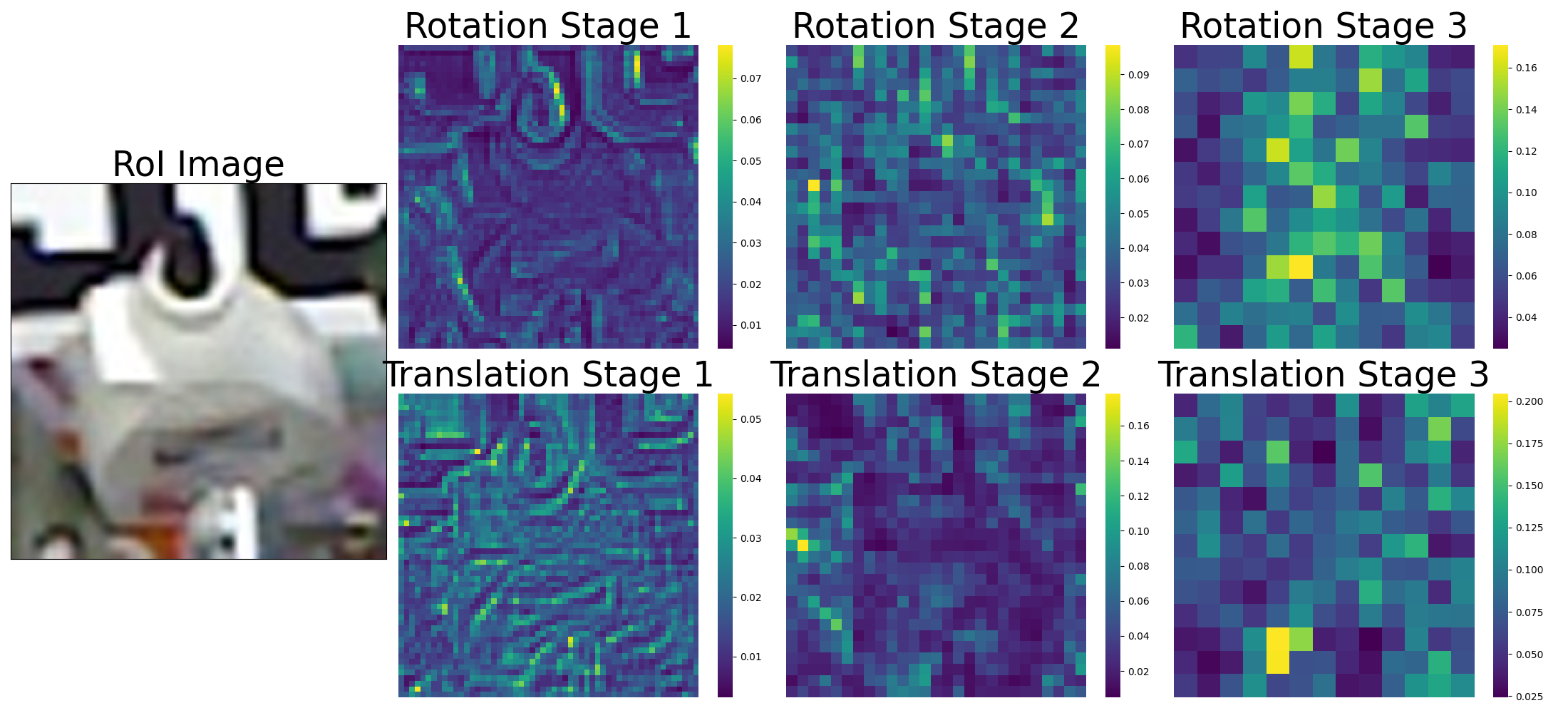}
    \end{subfigure}
    \vspace{10mm}

    \begin{subfigure}{0.48\textwidth}
        \centering
        \includegraphics[width=\linewidth]{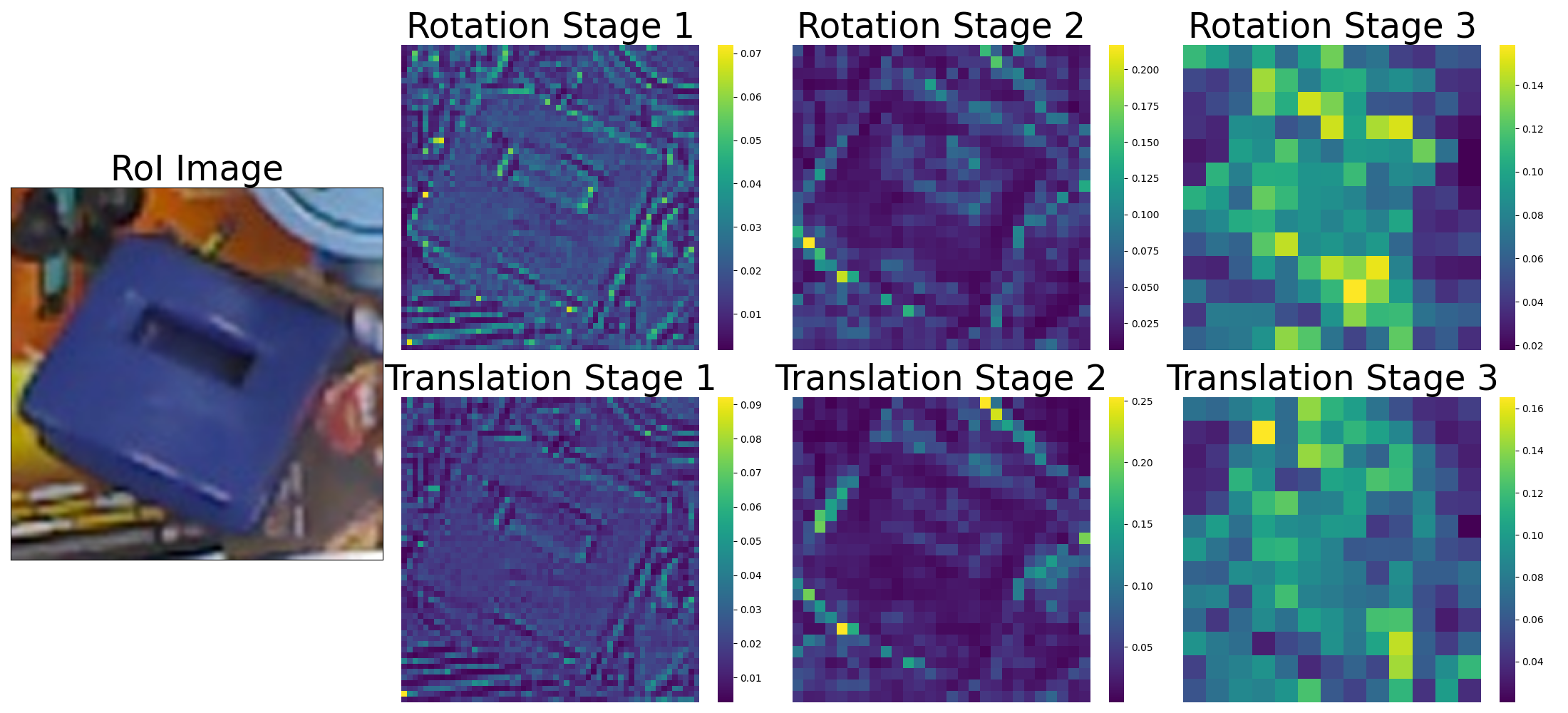}
    \end{subfigure}%
    \hfill
    \begin{subfigure}{0.48\textwidth}
        \centering
        \includegraphics[width=\linewidth]{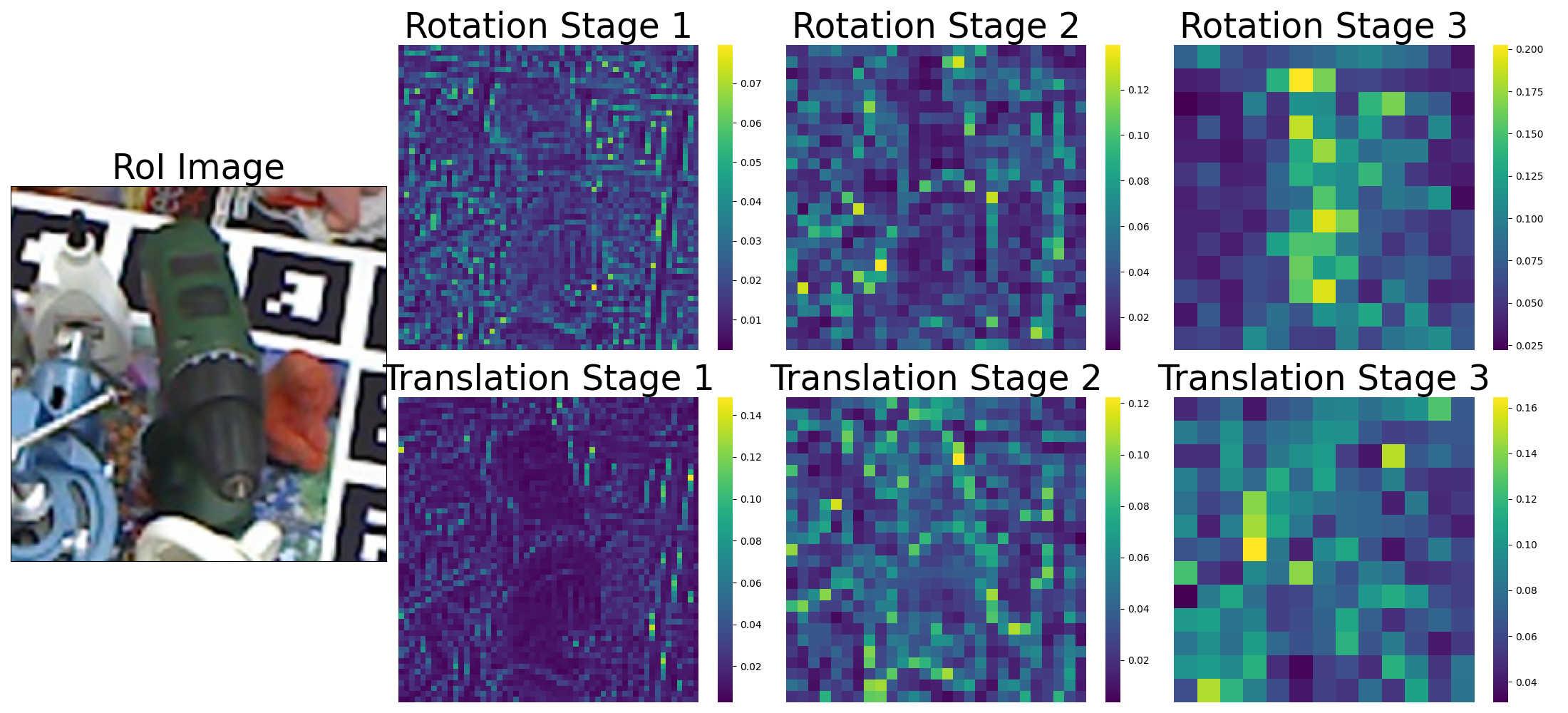}
    \end{subfigure}%
    \vspace{10mm}

    \begin{subfigure}{0.48\textwidth}
        \centering
        \includegraphics[width=\linewidth]{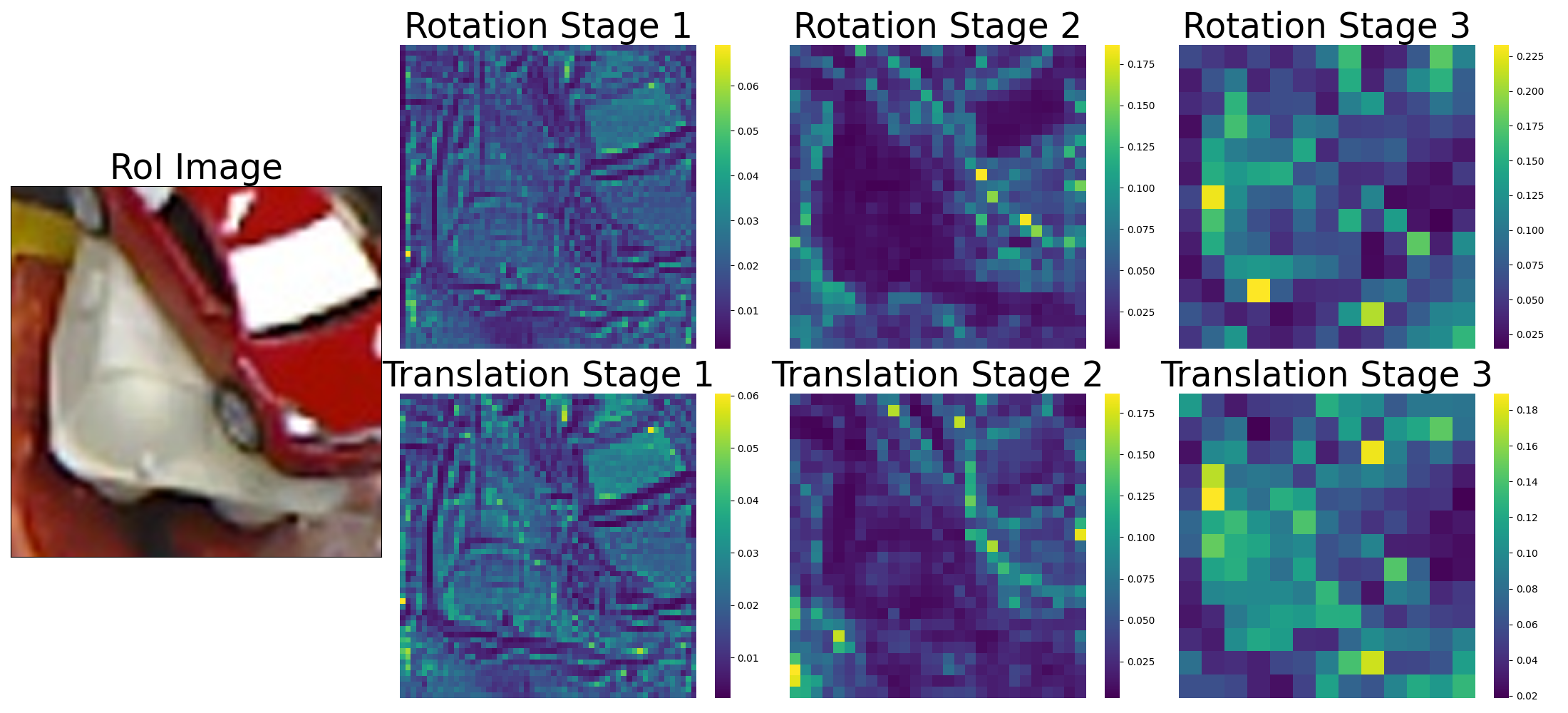}
    \end{subfigure}%
    \hfill
    \begin{subfigure}{0.48\textwidth}
        \centering
        \includegraphics[width=\linewidth]{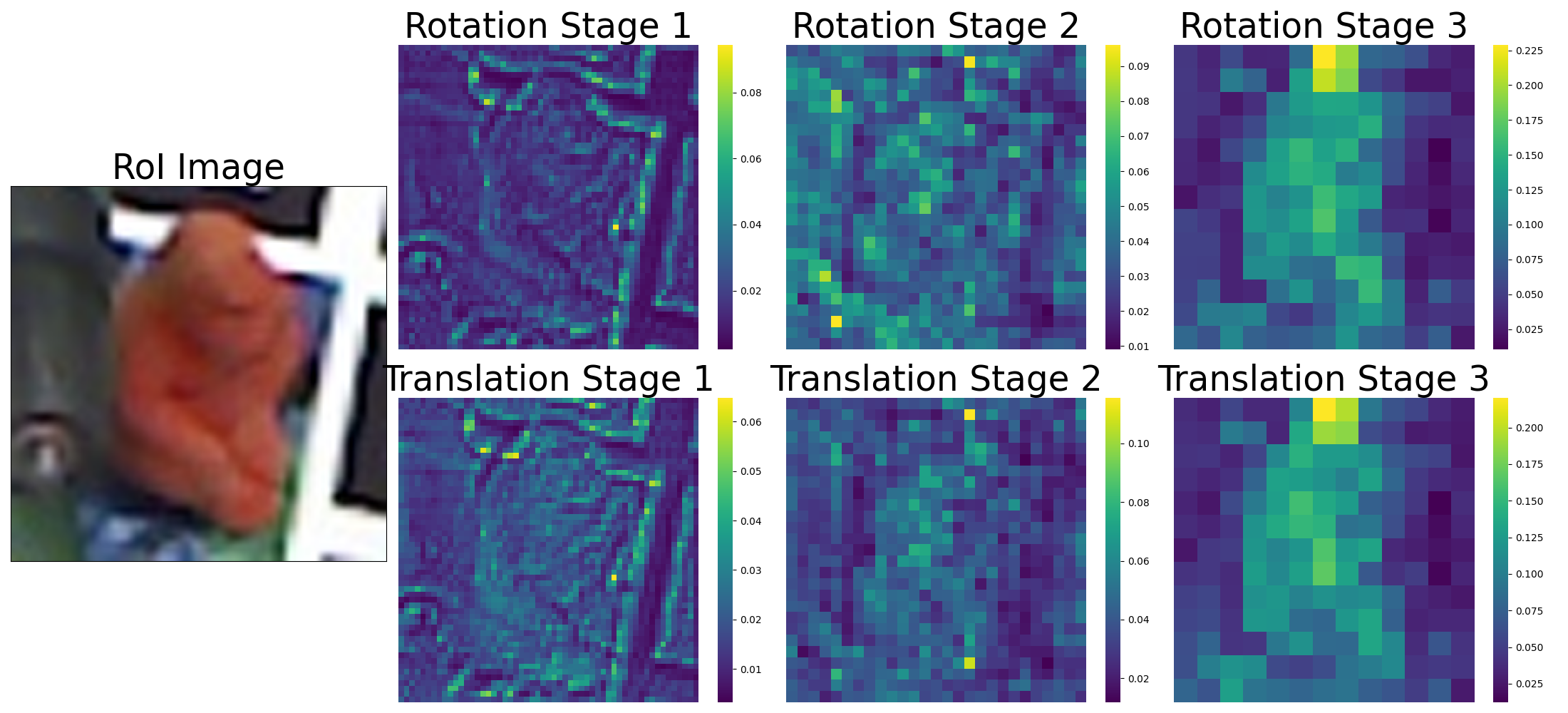}
    \end{subfigure}%
    \caption{\textbf{Attention maps for different objects of the LM-O dataset.} The top row depicts the rotation attention,
        and the bottom row shows the translation attention. The columns depict the different stages of the
        transformer.}
    \label{fig:test}
\end{figure*}

Further examples of attention maps for different objects are shown in
\cref{fig:test}. The first column shows the sample RoI image. The first row shows the rotation attention map, 
and the second row shows
the translation attention map. The columns depict the different stages
of the model. We visualize one random example from the
LM-O dataset for each object.

\section{Late Fusion Saliency Map}
\begin{figure*}[ht]
    \centering
    \includegraphics[width=0.8\textwidth]{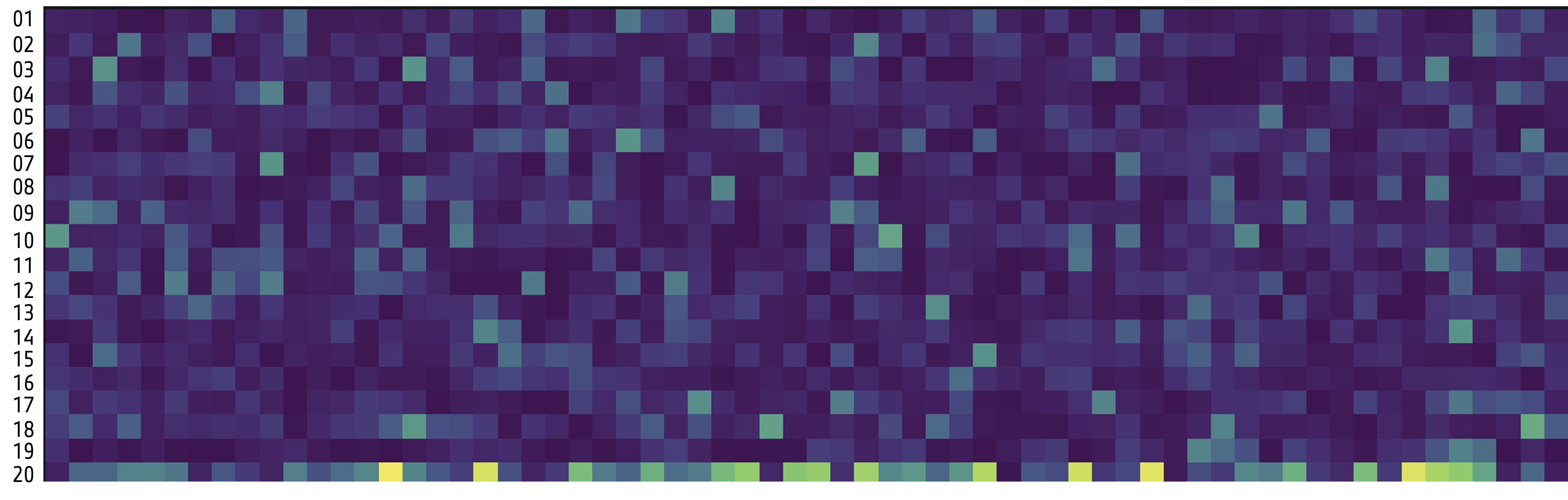}
    \caption{\textbf{Saliency Map of Pooling-Version.}
        The saliency map for the late fusion of class and bounding box
        embeddings is structured as follows: The initial 18 rows depict the
        1152-dimensional pooled feature vector. The second last row represents the
        bounding box embedding, and the final row is dedicated to the class embedding.
        This visualization is based on the mean values derived from a batch of 1000 random samples.}
    \label{tab:late_fusion}
\end{figure*}

When using the Pool version, see \cref{fig:tokens}, we appended the class and bounding box embedding
to the pooled feature map.
To demonstrate the impact of this modification, we present a saliency map of this feature vector, focusing on its relation to the pose loss.
As indicated in \cref{tab:late_fusion}, we observe that the saliency of
the second last row, the bounding box embedding, is significantly lower compared to the
feature vector and the class embedding. This outcome aligns with our expectations
as
the pose representation is designed to be unaffected by variations in the bounding
box. Conversely, the class embedding plays a crucial role in the prediction process,
as indicated by the high saliency in the last row.

